\documentclass[sigconf]{acmart}
\AtBeginDocument{%
	\providecommand\BibTeX{{%
			\normalfont B\kern-0.5em{\scshape i\kern-0.25em b}\kern-0.8em\TeX}}}

\copyrightyear{2022} 
\acmYear{2022} 
\setcopyright{acmcopyright}\acmConference[CIKM '22]{Proceedings of the 31st ACM International Conference on Information and Knowledge Management}{October 17--21, 2022}{Atlanta, GA, USA}
\acmBooktitle{Proceedings of the 31st ACM International Conference on Information and Knowledge Management (CIKM '22), October 17--21, 2022, Atlanta, GA, USA}
\acmPrice{15.00}
\acmDOI{10.1145/3511808.3557103}
\acmISBN{978-1-4503-9236-5/22/10}

\usepackage[ruled]{algorithm2e}
\usepackage{subfigure}
\usepackage{stfloats}
\usepackage{multirow}
\usepackage{color}
\usepackage[ruled]{algorithm2e}
\usepackage{enumitem}
\usepackage{balance}

\begin{document}
	
\title{Hierarchically Constrained Adaptive Ad Exposure
	in Feeds}

\author{Dagui Chen$^{*}$}
\affiliation{
	\institution{Alibaba Group}
	\country{}
}
\email{dagui.cdg@alibaba-inc.com}

\author{Qi Yan$^{*}$}
\affiliation{
	\institution{Alibaba Group}
	\country{}
}
\email{qiyan.yq@alibaba-inc.com}

\author{Chunjie Chen}
\affiliation{
	\institution{Alibaba Group}
	\country{}
}
\email{chunjie.ccj@alibaba-inc.com}

\author{Zhenzhe Zheng$^{\dagger}$}
\affiliation{
	\institution{Shanghai Jiao Tong University}
	\country{}
}
\email{zhengzhenzhe@sjtu.edu.cn}

\author{Yangsu Liu}
\affiliation{
	\institution{Shanghai Jiao Tong University}
	\country{}
}
\email{liu_yangsu@sjtu.edu.cn}

\author{Zhenjia Ma}
\affiliation{
	\institution{Alibaba Group}
	\country{}
}
\email{mazhenjia.mzj@alibaba-inc.com}

\author{Chuan Yu}
\affiliation{
	\institution{Alibaba Group}
	\country{}
}
\email{yuchuan.yc@alibaba-inc.com}

\author{Jian Xu}
\affiliation{
	\institution{Alibaba Group}
	\country{}
}
\email{xiyu.xj@alibaba-inc.com}

\author{Bo Zheng}
\affiliation{
	\institution{Alibaba Group}
	\country{}
}
\email{bozheng@alibaba-inc.com}

\renewcommand{\shortauthors}{Dagui Chen et al.}
\renewcommand{\shorttitle}{Hierarchically Constrained Adaptive Ad Exposure in Feeds}

\begin{abstract}
	A contemporary feed application usually provides blended results of organic items and sponsored items~(ads) to users.
	Conventionally, ads are exposed at fixed positions.
	Such a fixed ad exposure strategy is inefficient due to ignoring users' personalized preferences towards ads.
	To this end, \textit{adaptive ad exposure} is becoming an appealing strategy to boost the overall performance of the feed. 
	However, existing approaches to implement the adaptive ad exposure strategy suffer from several limitations: 
	1) they usually fall into sub-optimal solutions because of only focusing on request-level optimization without consideration of the application-level performance and constraints,
	2) they neglect the necessity of keeping the game-theoretical properties of ad auctions,
	and 3) they can hardly be deployed in large-scale applications due to high computational complexity.
	In this paper, we focus on the application-level performance optimization under hierarchical constraints in feeds and formulate adaptive ad exposure as a Dynamic Knapsack Problem.
	We propose Hierarchically Constrained Adaptive Ad Exposure~(HCA2E) that possesses the desirable game-theoretical properties, computational efficiency, and performance robustness.
	Comprehensive offline and online experiments on a leading e-commerce application demonstrate the performance superiority of HCA2E.
	{\let\thefootnote\relax\footnote{
			{$^{*}$Equal contribution. $^{\dagger}$Corresponding author.}}}
\end{abstract}



\maketitle

\section{Introduction}

\begin{figure}[t] 
	\centering
	\includegraphics[width=0.95\linewidth]{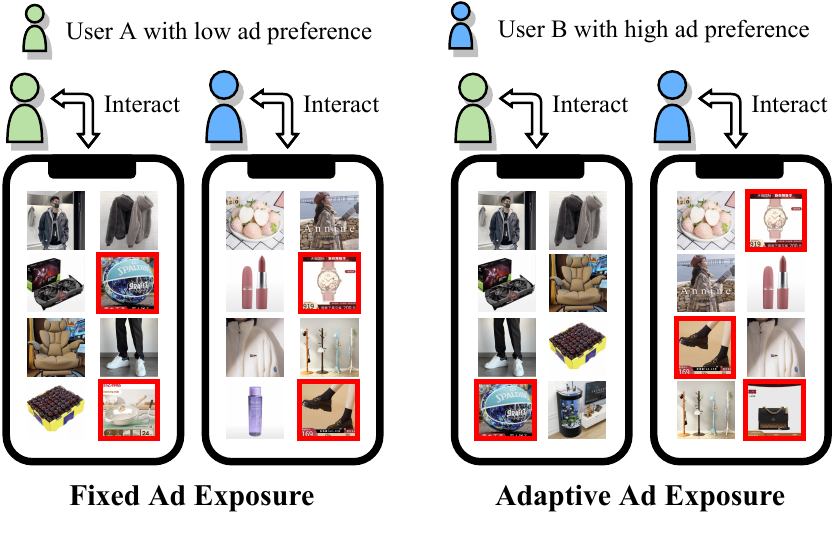}
	\caption{Fixed/adaptive ad exposure in e-commerce feed. The adaptive strategy can dynamically determine ad positions (red boxes) according to the user preference for ads.
	}
	\Description{}
	\label{fig:ecom}
\end{figure}

Nowadays, many online applications such as e-commerce, news recommendation, and social networks organize their contents in feeds.
A contemporary feed application usually provides blended results of organic items~(recommendations) and sponsored items~(ads) to users~\cite{chen2019learning}.
Conventionally, ad exposure positions are fixed for the sake of simplicity in system implementation, which is known as \emph{fixed ad exposure} (left part in \figurename~\ref{fig:ecom}). 
However, such a fixed exposure strategy can be inefficient due to ignoring users' personalized preferences towards ads and lacks the flexibility to handle changes in business constraints~\cite{geyik2016joint, zhang2021optimizing}.
For example, there are more active users during e-commerce sale promotions, and thus exposing more ads than usual time could be more favorable.
To this end, \textit{adaptive ad exposure}~(right part in \figurename~\ref{fig:ecom}), dynamically determining the number and positions of ads for different users, has become an appealing strategy to boost the performance of the feed and meet various business constraints.

However, there are several critical challenges in implementing the adaptive ad exposure strategy in large-scale feed applications.
\textbf{First}, adaptive ad exposure is a multi-objective optimization problem.
For example, it needs to strike a balance between recommendation-side user engagement and advertising-side revenue,
because more ads will generally increase ad revenue at the cost of user engagement.
Thus, the performance of feeds with adaptive ad exposure should be Pareto efficient~\cite{lin2019pareto}.
\textbf{Second}, {business constraints, usually consisting of both request-level constraints~(for a single user request) and application-level constraints~(for all the requests over a period), should be taken into account due to the business nature.}
At the fine-grained request level, ad positions are constrained (e.g., not too dense) for a good user experience \cite{yan2020ads}.
At the application level, the \textit{monetization rate}, indicating the average proportion of ad exposures, should be constrained by an upper bound~\cite{wang2019learning}.
It is difficult to optimize the adaptive ad exposure strategy under the entanglement of hierarchical constraints.
\textbf{Third}, some desirable game-theoretical properties of ad auctions should be guaranteed. {For example, Incentive Compatibility~(IC)~\cite{vickrey1961counterspeculation} and Individual Rationality~(IR)~\cite{nazerzadeh2013dynamic} 
theoretically guarantee that truthful bidding is optimal for each advertiser, which is important for the long-term prosperity of the ad ecosystem~\cite{aggarwal2009general, wilkens2017gsp, liu2021neural}. }
\textbf{Fourth}, the adaptive ad exposure strategy must be computationally efficient to achieve low-latency responses
and be robust enough to guarantee the performance stability in the ever-changing online environment~\cite{yan2020ads}.

A few efforts have been made to study adaptive ad exposure. 
Lightweight \emph{rule-based} algorithms~\cite{wang2011learning, zhang2018whole, yan2020ads} are designed to blend organic items and ads following a ranking rule with some predefined re-ranking scores. 
However, they focus on the request-level optimization without consideration of application-level performance,
resulting in sub-optimal performance of the feed applications under hierarchical constraints.
\emph{Learning-based} methods~\cite{wang2019learning, zhao2020jointly, zhao2021dear, liao2021cross} mostly employ reinforcement learning~(RL)~\cite{sutton2018reinforcement} to search for the optimal strategies and perform well in offline simulations.
However, since RL models heavily rely on massive training data to update the parameters,
they are neither robust nor lightweight to deal with the rapidly changing online environment.
We also note that most works leave out the discussion of the necessity to guarantee desirable game-theoretical properties.

In this work, towards improving application-level performance under hierarchical constraints, we formulate adaptive ad exposure as a Dynamic Knapsack Problem~(DKP)~\cite{dizdar2011revenue, hao2020dynamic} and propose a new approach Hierarchically Constrained Adaptive Ad Exposure~(\textbf{HCA2E}).
More specifically, to alleviate the difficulties in handling the hierarchical constraints,
we design a two-level optimization architecture that decouples the DKP into request-level optimization and application-level optimization.
In request-level optimization, we introduce a Rank-Preserving Principle~(RPP) to maintain the game-theoretical properties of the ad auction and propose an Exposure Template Search~(ETS) algorithm to search for the optimal exposure result satisfying the request-level constraints.
In application-level optimization, we employ a real-time feedback controller to adapt to the application-level constraint.
Moreover, we demonstrate that the two-level optimization is computationally efficient and robust against online fluctuations.
{Finally, comprehensive offline and online experimental evaluation results demonstrate that HCA2E outperforms competitive baselines.}

The main contributions of this paper are summarized as follows:
\begin{itemize}[leftmargin=*]
	\item  We formulate adaptive ad exposure in feeds as a Dynamic Knapsack Problem~(DKP), which can capture various business optimization objectives and hierarchical constraints to derive flexible Pareto-efficient solutions.
	\item We provide an effective solution: Hierarchically Constrained Adaptive Ad Exposure~(HCA2E), which possesses desirable game-theoretical properties, computational efficiency, and performance robustness.
	\item We have deployed HCA2E on Taobao, a leading e-commerce application. Extensive offline simulation and online A/B experiments demonstrate the performance superiority of HCA2E over competitive baselines.
\end{itemize}

\begin{figure}[tp] 
	\centering
	\includegraphics[width=0.95\linewidth]{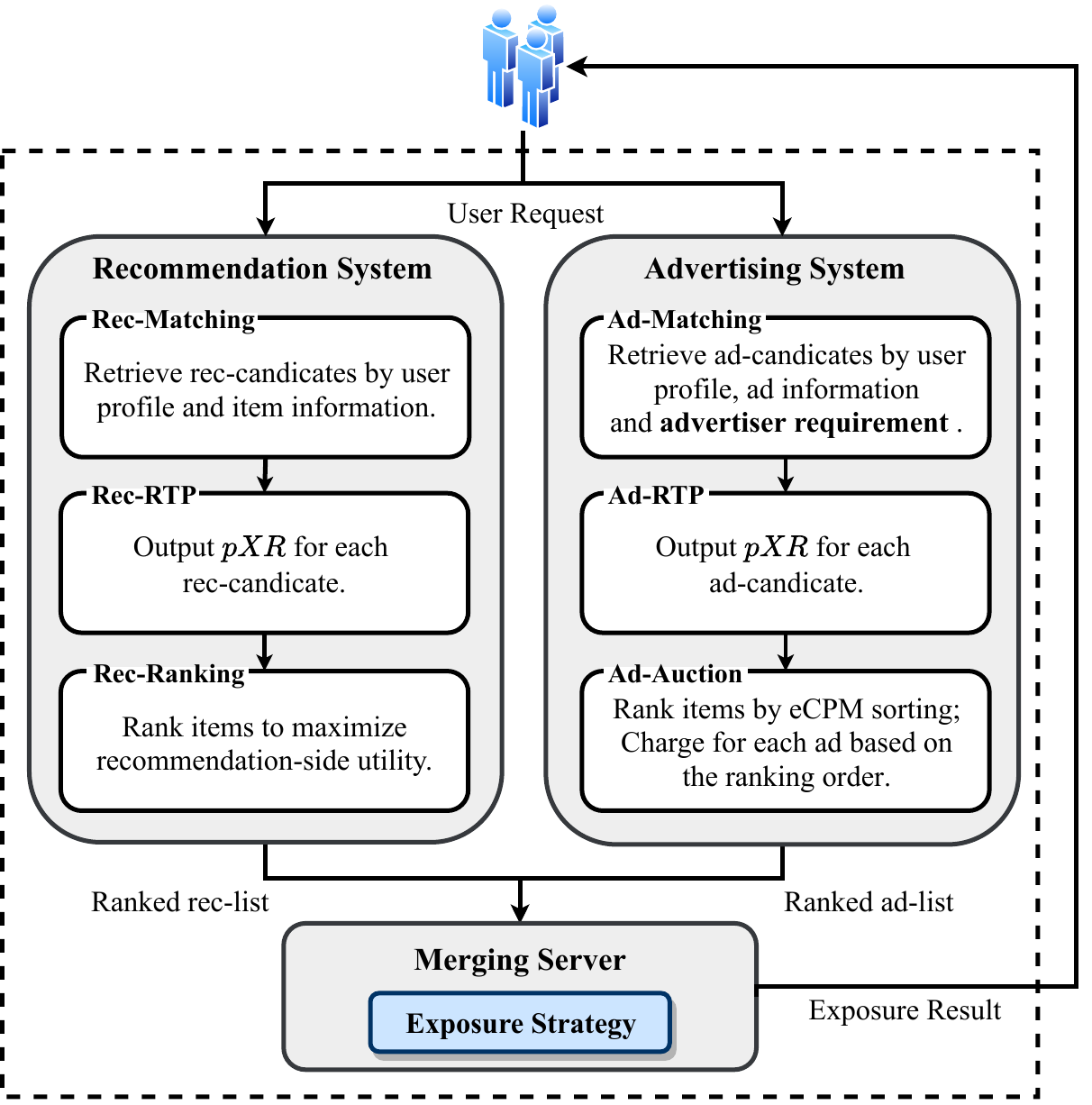}
	\caption{The overview of a typical feed system.}
	\Description{The illustration of e-commerce feed framework. 
		When a request arrives, Recommendation System~(RS) and Advertising System~(AS) generate ranked lists of recommended and advertised candidates in parallel. Merging Server~(MS) aggregates them as a hybrid result to display.}
	\label{fig:arch}
\end{figure}

\section{Background} \label{sec:back}
In this section, we first present a system overview of a typical feed application.
Then, we make necessary descriptions of major objectives and constraints of the performance optimization.
We note that the introduced system framework, objectives, and constraints are applicable in various feed applications.

\subsection{Feed Application System}

As illustrated in Figure \ref{fig:arch}, a typical feed application has three parts:
\begin{itemize}[leftmargin=*]
\item
{\textbf{Recommendation System~(RS). }}
The running of RS consists of three cascading phases: 
\textbf{1)} Rec-Matching server retrieves hundreds of candidate items according to the relevance between users' preferences and items.
\textbf{2)} Rec-RTP server predicts multiple performance indicators for each candidate, such as $ pXR $~(short for predicted $ X $ rate, where $ X $ can be click-through, favorite, conversion, etc.).
\textbf{3)} Rec-Ranking server sorts candidate items to maximize the expected recommendation utility, such as user engagement in social networks and gross merchandise volume in e-commerce feeds.
\item
{\textbf{Advertising System~(AS). }}
AS has three similar phases to RS.
However, due to the advertisers' participation, several differences exist:
\textbf{1)}
Besides user-item relevance, {Ad-Matching} server needs to take advertisers' willingness (e.g. bids and budgets) into account~\cite{jin2018real}.
\textbf{2)} 
Different from the Rec-Ranking server, {Ad-Auction} server should follow a certain auction mechanism, including the rules of ad allocation and pricing. 
With the predicted performance indicators, the ad allocation rule ranks candidate ads by descending order of \emph{effective Cost Per Mille}~(eCPM) to maximize the expected ad revenue~\cite{zhu2017optimized}.
The pricing rule determines the payment for the winning ads, such as the critical bid in the Generalized Second Price~(GSP) auction~\cite{wilkens2017gsp}.
\item
{\textbf{Merging Server~(MS). }}
Given two ranked lists of recommendation and ad candidates, MS combines them according to an exposure strategy that determines the order of exposed items for each request.
Corresponding to the business requirements of the application, MS is desirable to achieve flexible Pareto-efficient solutions to trade off recommendation-side and advertising-side objectives.
In addition, the computational complexity of MS should be enough low to ensure low latency.
\end{itemize}

\subsection{Performance Objectives}  \label{sec:utility}
The major performance objectives of the feed application contain {\textbf{recommendation utility}} and {\textbf{advertising utility}}.
Recommendation utility (from recommendations) could have different interpretations in different feed applications.
For example, user engagement, driven by users' effective activities~(e.g., clicks, favorites, comments, and shares), is widely used in social networks and news recommendations.
In e-commerce feeds, gross merchandise volume~(GMV), relevant to users' clicks and further conversions, is the key indicator to measure the total amount of purchased commodities over a period.
Advertising utility includes not only the above performance indicators (from ads) but also ad revenue.
The advertising revenue is a common metric to represent the total payment of exposed or clicked ads over a specific period.

\subsection{Business Constraints} \label{sec:constraint}
Without loss of generality, in this paper, we consider the following crucial business constraints:
\begin{itemize}[leftmargin=*]
\item
\textbf{Application-level Constraint.}
At the application level, the constraint of \textit{monetization rate} is concerned.
The {monetization rate} represents the proportion of ad exposures over total exposures (both recommendations and ads)  within a period. 
As a key metric to quantify the supply of ad resources, the monetization rate should be constrained by a target value for two reasons:
\textbf{1)}~fluctuating supply of ad resources might cause instability to the auction environment;
\textbf{2)}~a too large proportion of ads could damage the recommendation utility.
\item
\textbf{Request-level Constraints. }
To prevent some unintended results from harming the user experience, the ad exposure strategy should be further restricted by some request-level constraints.
Similar to \cite{yan2020ads}, the following two constraints are considered:
\textbf{1)} \textit{top ad slot} (TAS) defines the allowed highest exposure position of ads among all items to be exposed; 
\textbf{2)} \textit{minimum ad gap} (MAG) defines the allowed minimum position distance between two adjacent ads.
\item
\textbf{Auction Mechanism Guardrails. }
The allocation and pricing of ads are based on certain auction mechanisms.
To satisfy desirable game-theoretical properties (e.g., IC and IR as aforementioned), the pricing rule of an auction mechanism should be highly correlated to the ranking results~\cite{myerson1981optimal}.
Thus, we need to introduce additional constraints to keep these properties,
because the merging process may change the original rankings.
\end{itemize}

\section{Problem Formulation}\label{sec:form}

Within a period~(e.g., one day), we consider all the user requests as a sequence $\mathcal{S}$.
For a request $ s \in \mathcal{S} $, an ad exposure strategy $ \pi_s $ determines the placement (positions and orders) of ads.
Under $ \pi_s $, the request-level expected utility  $ U(s|\pi_s) $ is determined by all items to be exposed on $s$, and should be a trade-off between expected rec-utility $ U^{\rm rec}(s|\pi_s) $ and ad-utility $ U^{\rm ad}(s|\pi_s) $, calculated as follows\footnote[1]{
	$ U^{\rm rec} $ and $ U^{\rm ad} $ are consistent with the ranking scores of RS and AS, and could be quantified by different performance metrics according to the business requirements.}:
\begin{align} \label{eq:util}
	U(s|\pi_s)  \triangleq U^{\rm ad}(s|\pi_s)  + \alpha \times  U^{\rm rec}(s|\pi_s),
\end{align} 
where $ \alpha $ is an adjustable trade-off hyper-parameter.
We optimize $ \pi $ by maximizing the accumulative utility over $\mathcal{S}$, and the optimization should be constrained as described in Section \ref{sec:constraint}.
At the application level, the monetization rate $m (\mathcal{S})$ should be constrained by a target value $ m^* $, i.e.,
\begin{align} \label{eq:pvr}
	m(\mathcal{S}) \triangleq \frac{\sum_{s \in \mathcal{S}}  N^{\rm ad}(s|\pi_s) }{\sum_{s \in \mathcal{S}} N(s|\pi_s)}  \leq m^*,
\end{align} 
where $ N^{\rm ad}(s) $ and $ N(s|\pi_s) $ denote the number of ads and all items respectively.
At the request level, ad positions are restricted by the top ad slot and minimum ad gap. 
Thus, the overall optimization problem could be formulated as:
\begin{align} \label{eq:org-obj}
	&\mathop{\rm maximize}_{\pi_s}  \sum_{s \in \mathcal{S} }  U(s|\pi_s)  \\
	\text{subject } &{\rm to  }
	\begin{cases}
		\sum_{s \in \mathcal{S}}N^{\rm ad}(s|\pi_s)  \leq m^* \times  \sum_{s \in \mathcal{S}} N(s|\pi_s) 
		\\
		\pi_s  \in  \Pi_{\rm TAS} \cap \Pi_{\rm MAG} \notag
	\end{cases},
\end{align} 
where we use $ \Pi_{\rm TAS} $ and $ \Pi_{\rm MAG} $ to denote the strategy space satisfying the request-level constraints of top ad slot (TAS) and minimum ad gap (MAG).

\begin{table}[!t] 
	\caption{The mainly used notations.}
	\label{tab:nota}
	\renewcommand\arraystretch{1.2}
	\setlength{\tabcolsep}{1.6mm}{
		\begin{tabular}{l | l}
			\toprule
			\hline
			Notation &  Description \\  
			\hline
			$ \mathcal{S}, s $  &  Request sequence and a single request. \\
			$ U^{\rm rec}, U^{\rm ad}, U $  &  
			\begin{tabular}[c]{@{}l} 
			Expected rec-utility, ad-utility, and \\ overall utility of a request. 
			\end{tabular} \\
			$ N^{\rm rec}, N^{\rm ad}, N $  &  
			\begin{tabular}[c]{@{}l} 
			Expected rec-exposures, ad-exposures,  \\ and total exposures of a request.
			\end{tabular} \\
			$ m, m^* $ &  Monetization rate and its target value. \\
			$ v $,$ w $ &  Value and weight of a request. \\
			$ V $, $W$, $W^*$  &  
			\begin{tabular}[c]{@{}l} 
			Accumulative value, accumulative weight  \\ and capacity of the knapsack.
			\end{tabular} \\
			$ \mathbf{x}, \pi $ &  Selection strategy and exposure strategy. \\
			$ \rho, \rho_{{}_{\rm THRES}} $ &  Value per weight and its threshold. \\
			$ \Delta v $ &  Knapsack value increment by a request. \\
			\hline
			\bottomrule
		\end{tabular}
	}
\end{table}

\section{Hierarchically Constrained Adaptive Ad Exposure} \label{sec:meth}

\subsection{Dynamic Knapsack Problem}

\subsubsection{\textbf{Strategy Decomposition}}
The optimization problem~(\ref{eq:org-obj}) is challenging due to several reasons:
\textbf{1)}~The optimization variable $ \pi $ indicating different ranking results is discrete, and thus the optimization objective could be non-convex.
\textbf{2)}~The optimization of $\pi$ is hierarchically constrained, i.e., real-time exposure on a single request needs to take into account the application-level constraint across all requests in $\mathcal{S}$.
\textbf{3)}~$\pi$ should also guarantee the auction mechanism properties and high computational efficiency.
Thus, it is hard to directly optimize $\pi$.
To tackle these challenges, we consider designing a hierarchical optimization approach, where we decompose the strategy over $\mathcal{S}$ into two levels:
\begin{itemize}[leftmargin=*]
	\item \textit{application-level selection strategy} $ \mathbf{x} = [ x_s ]_{s \in \mathcal{S} }$ where $ x_s = 1 $ as $s$ is selected to expose ads and $ x_s = 0 $ otherwise. 
	\item \textit{request-level exposure strategy} $ \pi $ determining how the candidate ads are exposed on a selected request.
\end{itemize}
Specially, we denote the exposure strategy without ads by $ \pi^0 $, and $ \pi_s = \pi^0$ if $ x_s = 0 $ given a request $ s $.
Hence, the optimization problem~(\ref{eq:org-obj}) could be considered as a Knapsack Problem \cite{salkin1975knapsack}.

\subsubsection{\textbf{Request Description}}
From the perspective of a Knapsack Problem, each request is treated as an object with individual value and weight.
For a request $s$ selected to expose ads, the real value produced by an exposure strategy $\pi_s$ should be the incremental utility of $ \pi_s $ to the no-ad strategy $ \pi^0 $.
Accordingly, we define the value of $ s $ under $\pi_s$ as:
\begin{align} \label{eq:value}
	v(s|\pi_s) \triangleq  U(s|\pi_s) - U(s|\pi^0).
\end{align}
Here, $ U(s|\pi^0) $ is fixed given request $s$ since the expected recommendation utility $ U^{\rm rec}(s|\pi^0) $ only depends on the recommended items ranked in RS.
Meanwhile, exposing ads could occupy a proportion of ad resources given the monetization rate constraint.
Then we consider the weight of $ s $ under $\pi_s$ to the knapsack, i.e.,
\begin{align} \label{eq:weight}
	w(s|\pi_s) \triangleq N^{\rm ad} (s|\pi_s),
\end{align} 
Specially, for $s$ under $\pi^0$, we have $ v(s|\pi^0) = 0$ and $ w(s|\pi^0)  = 0$.

\subsubsection{\textbf{Knapsack Description}}
Next, we consider a knapsack to accommodate the requests under the capacity constraint.
The knapsack value $V$ is defined as the total value of requests selected in the knapsack, i.e.,
$
	V \triangleq \sum_{s \in \mathcal{S} }  x_s \times v(s|\pi_s).
$ 
According to Equation (\ref{eq:pvr}), accumulative request weight in the knapsack (i.e., total ad exposures within $ \mathcal{S} $) should not exceed an upper bound, i.e., $ \sum_{s \in \mathcal{S}} N^{\rm ad}(s|\pi_s) \leq  m^* \times \sum_{s \in \mathcal{S}} N(s|\pi_s) $.
The accumulative request weight $ W \triangleq \sum_{s \in \mathcal{S}} N^{\rm ad}(s|\pi_s) $, and the knapsack capacity could be expressed as
\begin{align} \label{eq:cap}
	W^* \triangleq m^*  \times \sum_{ s \in \mathcal{S} } N(s|\pi_s).
\end{align} 
In this work, we consider that $W^*$ is constant within $\mathcal{S}$ because the target monetization rate is predetermined and $ \pi $ could rarely affect total exposures within a specific period.

\subsubsection{\textbf{Optimization Objective}}
Based on the above descriptions, the goal is to maximize the knapsack value under the capacity constraint and aforementioned request-level constraints. 
The optimization problem (\ref{eq:org-obj}) could be further expressed as:
\begin{align} \label{eq:obj}
	&\mathop{\rm maximize}\limits_{\mathbf{x}, \pi_s \neq \pi^0}  \sum_{s \in \mathcal{S} }  x_s \times v(s|\pi_s) \\
	&\text{subject }\text{to  }
	\begin{cases}
		\sum\limits_{ s \in \mathcal{S} }  x_s \times w(s|\pi_s)  \leq  W^*  \\
		\pi_s  \in  \Pi_{\rm TAS} \cap \Pi_{\rm MAG} \notag
	\end{cases}.
\end{align} 
Different from a classic knapsack problem \cite{salkin1975knapsack}, since both the value and weight of each object (request) depend on $ \pi $ and thus are variable, Formulation (\ref{eq:obj}) is a Dynamic Knapsack Problem (DKP).

We handle the DKP by the following steps: 
\textbf{1)} 
We design a two-level optimization architecture, consisting of application-level optimization and request-level optimization.
\textbf{2)}
To maintain the game-theoretical properties and the independence/flexibility of AS and RS, we preserve the prior order from AS/RS, termed as Ranking-Preserving Principle, which also reduces the optimization complexity.
\textbf{3)}
Based on the above designs, we use two lightweight algorithms to effectively solve the application-level optimization and request-level optimization respectively.
We summarize our solutions as a unified approach: Hierarchically Constrained Adaptive Ad Exposure~(HCA2E).
We also reveal the implementation details of HCA2E for large-scale online services.

\subsection{Hierarchical Optimization Formulation}\label{sec:kvim}

\subsubsection{\textbf{Application-level Optimization}}\label{sec:greedy}

\begin{figure}[tp] 
	\centering
	\includegraphics[width=0.95\linewidth]{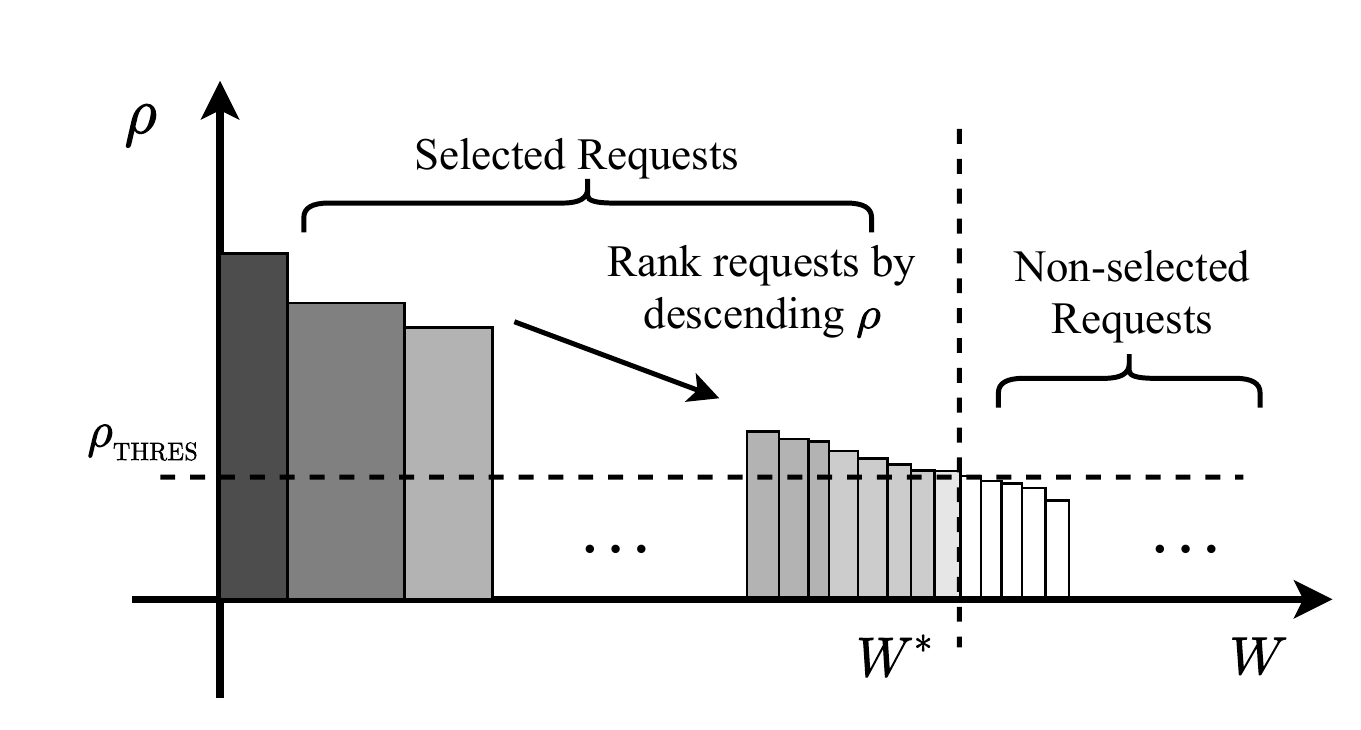}
	\caption{Greedy algorithm for optimizing $\mathbf{x}$.}
	\label{fig:vwr}
\end{figure}

We first consider optimizing the application-level selection strategy $ \mathbf{x} $ under the determined $ \pi  $, i.e., $ v(s|\pi_s) $ and $ w(s|\pi_s) $ of any $ s \in \mathcal{S} $ are determined. 
The Greedy algorithm~\cite{dantzig1955discrete} has been proposed to solve such a static 0-1 Knapsack Problem, where \textit{{Value Per Weight}} (VPW) is calculated by:
\begin{align} \label{eq:vwr}
	\rho (s | \pi_s ) = \frac{v(s | \pi_s )}{w( s|\pi_s )}.
\end{align}
Specially, we note that $ \rho (s | \pi^0 ) = 0 $ as $ v (s | \pi^0 ) = 0 $.  
The basic idea is to greedily select the requests by descending order of $\rho$.
As illustrated in Figure \ref{fig:vwr}, we rank the requests and select the top requests until the cumulative weight exceeds the capacity $ W^* $.

However, ranking all the requests could be impractical in large-scale industrial scenarios.
From Figure \ref{fig:vwr}, we note that the selected requests depend on a screening threshold $ \rho_{{}_{\rm THRES}} $:
the requests with larger $ \rho $ than $ \rho_{{}_{\rm THRES}} $ will be selected into the knapsack.
If $ \rho_{{}_{\rm THRES}} $ is determined, we can obtain $ {x}_s $ for $ s  \in \mathcal{S} $ as: 
\begin{align}\label{eq:greedy}
	x_s = \mathbb{I} \left[ \rho(s|\pi_s) > \rho_{{}_{\rm THRES}} \right],
\end{align}
where $ \mathbb{I}[\ast] $ is the indicator function.
Thus, the optimization of $\mathbf{x}$ can be transformed as determining the 1-dimension variable $ \rho_{{}_{\rm THRES}} $.

\subsubsection{\textbf{Request-level Optimization}}

\begin{figure}[t] 
	\centering
	\includegraphics[width=0.95\linewidth]{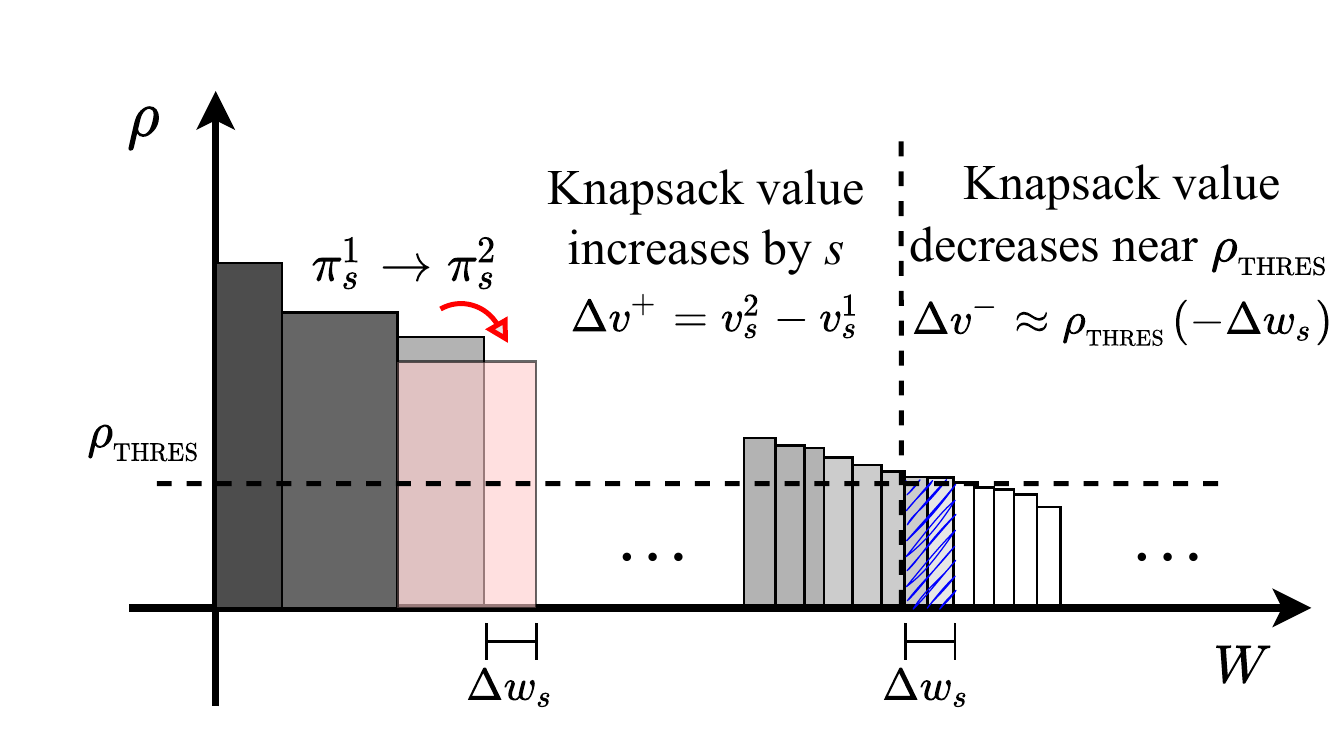}
	\caption{An illustration for designing request-level exposure strategy. From $ \pi^1_s $  to $ \pi^2_s $ with $ \Delta w_s > 0 $, if $ \pi^2_s $ outperforms $ \pi^1_s $, we should have $  \Delta v^{+} +  \Delta v^{-} > 0 $.}
	\label{fig:kvim}
\end{figure}

We then fix $\mathbf{x} $ to optimize the request-level exposure strategy $\pi$.
For a selected request $ s $, we consider two different ad exposure strategies $\pi^1_s$ and $\pi^2_s$.
As illustrated in Figure \ref{fig:kvim}, if the exposure is changed from $\pi^1_s$ to $\pi^2_s$, $ s $ will have a request-level value change, i.e., $ \Delta v^{+}= v(s|\pi^2_s) - v(s|\pi^1_s) $, and a request-level weight change, i.e., $ \Delta w_s = w(s|\pi^2_s) - w(s|\pi^1_s) $. 
Since requests are selected in the descending order of $\rho$ greedily, if the request weight changes, some requests near $ \rho_{{}_{\rm THRES}} $~(represented by the blue shaded area in Figure~\ref{fig:kvim}) would be squeezed out~(if $ \Delta w_s > 0 $) or selected~(if $ \Delta w_s < 0 $) to satisfy the capacity constraint.
Accordingly, the knapsack value change resulted by these requests is $ \Delta v^{-} \approx \rho_{{}_{\rm THRES}} \times ( -\Delta w_s ) $.
Thus, if $\pi^2_s$ outperforms $\pi^1_s$ with increasing knapsack value, we then have $ \Delta v^{+}+\Delta v^{-} > 0 $,~i.e.,
\begin{align} \label{eq:impr}
	v(s|\pi^2_s) - \rho_{{}_{\rm THRES}} \times  w(s|\pi^2_s) > v(s|\pi^1_s) - \rho_{{}_{\rm THRES}} \times w(s|\pi^1_s) .
\end{align}
According to the strategy improvement condition~(\ref{eq:impr}), we can acquire the optimal ad exposure strategy with a given $ \rho_{{}_{\rm THRES}} $ as:
\begin{align} \label{eq:kvim}
	\pi^*_s = \arg \max_{\pi_s} \left[ v(s|\pi_s) - \rho_{{}_{\rm THRES}}  \times w(s|\pi_s) \right],
\end{align}
where we define \textit{knapsack value increment} of $ s $ as 
\begin{align}\label{eq:kvi}
	\Delta v(s|\pi_s) \triangleq v(s|\pi_s) - \rho_{{}_{\rm THRES}}  \times w(s|\pi_s).
\end{align}

Combining with formulas~(\ref{eq:greedy}) and~(\ref{eq:kvim}),  we can express the final exposure strategy for each request as:
\begin{equation} \label{eq:comb}
	\pi_s =
	\begin{cases}
		\pi_s^*, &\text{if \ \ } \rho(s|\pi_s^*) > \rho_{{}_{\rm THRES}}; \\
		\pi^0, & \text{otherwise}.
	\end{cases}
\end{equation}
The detailed optimization methods of $\pi$ and $\rho_{{}_{\rm THRES}}$ will be described in Section \ref{sec: temp} and Section \ref{sec:pid} respectively.

\subsection{Rank-Preserving Principle}
Next, we aim to guarantee the optimization of $\pi$ under the ad auction mechanism.
In this work, we focus on the two major game-theoretical properties: Incentive Compatibility~(IC) and Individual Rationality~(IR).
Formally, if every participant advertiser bids truthfully (i.e., bids the maximum willing-to-pay price~\cite{liu2021neural}), IR will guarantee their non-negative profits, and IC can further ensure that they earn the best outcomes.
According to Myerson's theorem~\cite{myerson1981optimal}, to satisfy these properties, the charge price of an ad should be related to its ranking in the ad list.
However, most of the existing works blend ads and recommended items without considering the fact that changed orders of ads should affect the charge prices.
These methods might lead to the advertisers' misreporting, which is against the long-term stable performance of the advertising system.
To this end, \textbf{Rank-Preserving Principle} (RPP) is introduced:
1) the relative orders of items/ads in MS should be consistent with the prior orders from RS/AS ~\cite{yan2020ads}, 
and 2) the final charge price for each exposed ad should be consistent with the price determined in AS.

More than acting as a guardrail for ad auctions, RPP contains some other benefits:

\textbf{1)} RPP allows RS and AS to run independently.
Since both RS and AS have quite complex structures and are driven by respective optimization objectives, different groups are responsible for RS and AS in a large-scale feed application.
Thus, RPP makes a clear boundary between the two subsystems and thus facilitates rapid technological upgrading within each subsystem.

\textbf{2)} RPP can effectively reduce the optimization complexity of request-level exposure strategy $\pi$.
Under RPP, optimizing $ \pi $ only needs to determine each slot for an ad or a recommended item, and then place the items/ads into the slots in their prior orders.
Thus, $\pi$ can be represented by an \textbf{exposure template},
defined as an one-hot vector $ \phi = [ i_1, i_2, \cdots, i_L ] $ to label the request's slots, where $ i_l = 1 $ indicates that the $ l $-th slot is for an ad and $ i_l = 0 $ otherwise, and we assume that there are $ L $ slots on a request.
Specially, we denote the no-ad template by $\phi^0$ (corresponding to $\pi^0$ described in Section~\ref{sec:form}), i.e., $\phi^0=[0, 0, \cdots]$.
We use $ K $ to denote the number of candidate items (including recommended candidates and ad candidates) for each request.
The exposure results without/with RPP are respectively $ \frac{K!}{ (K-L)!}  $ and $ 2^L $, where $ \frac{K!}{ (K-L)!} \gg 2^L  $ since the number of candidates is usually much larger than the number of slots.
Thus, the complexity of optimizing $\pi$ is reduced.

\subsection{Request-level Optimization with Exposure Template Search} \label{sec: temp}

Based on RPP, we optimize $\pi$ by searching the optimal exposure template $\phi^*$ on each request. 
We denote the set of all possible templates by $\Phi$. The length of each template is $ L $.

\subsubsection{\textbf{Template Evaluation}}

Given a request $ s $, the optimal exposure strategy follows the equation~(\ref{eq:kvim}), and thus any template $ \phi \in \Phi $ should be evaluated by:
\begin{align} \label{eq:kvit}
	\Delta v(s|\phi)  =  v(s|\phi) - \rho_{{}_{\rm THRES}} \times w(s|\phi).
\end{align}
where the value $ v(s|\phi) $ and weight $ w(s|\phi) $ under $\phi$ are required.
We calculate $ v(s|\phi) $ and  $ w(s|\phi) $ by the sum of discounted utility and weight of the $ L $ items inserted into $\phi$.
Here, we consider an exposure possibility for each slot $ l $ ($ 1 \leq l \leq L $), and the exposure possibility is non-increasing as $ l $ increases since users explore the application from top to bottom.
Thus, for the item at the $l$-th slot, a discounted utility should be the product of the slot exposure possibility and its original utility, and a discounted weight should be the slot exposure possibility if an ad is located at the $ l $-th slot or $ 0 $ otherwise.

\begin{figure}[t] 
	\centering
	\includegraphics[width=0.95\linewidth]{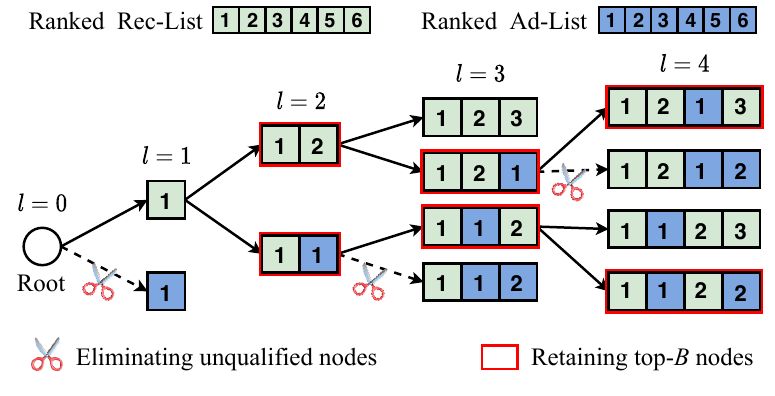}
	\caption{An illustration of Exposure Template Search. Here we set request length $ L = 4$, beam size $ B = 2$, top ad slot $ l_{\rm top} = 2$, and min ad gap $ \Delta l_{\rm min}=2 $.} 
	\label{fig:beam}
\end{figure}

\begin{algorithm}[t] 
	\caption{Exposure Template Search}
	\label{ag:beam}
	
	\LinesNumbered 
	\KwIn{user request $s$, request length $ L $, beam size $ B $, the VPW threshold $\rho_{{}_{\rm THRES}}$;}
	\KwOut{the final exposure template $\phi_s$;}
	Initialize a candidate template set $ \Phi_0 = \{\left[\right] \} $\; 
	\For{$ l = 1 : L $}{
		Initialize the $ l $-th layer's node buffer $ \Phi_l = \{ \} $\;
		\ForEach{ sub-template $ \kappa \in \Phi_{l-1} $ }{
			Expand $ \kappa $ with a rec-slot, i.e., $\kappa^{\rm rec} = \text{concat}(\kappa, [0])$\;
			Calculate $\Delta v(s|\kappa^{\rm rec})$ according to~(\ref{eq:kvit})\;
			Append $ \kappa^{\rm rec} $ into $ \Phi_l $, i.e., $ \Phi_l = \Phi_l \cup \{ \kappa^{\rm rec} \} $\;
			Expand $ \kappa $ with an ad-slot, i.e., $ \kappa^{\rm ad} = \text{concat}(\kappa, [1]) $\;
			Calculate $\Delta v(s|\kappa^{\rm ad})$ according to~(\ref{eq:kvit})\;
			Append $ \kappa^{\rm ad} $ into $ \Phi_l $, i.e., $ \Phi_l = \Phi_l \cup \{ \kappa^{\rm ad} \} $\;
		}
		Eliminate the nodes violating request-level constraints\;
		Rank the nodes in $\Phi_l $  by descending $ \Delta v(s|\kappa')_{\kappa' \in \Phi_l}$  \;
		Retain the top-$ B $ nodes, i.e., $ \Phi_l = \Phi_l [:B] $;  
	}
	\Return $ \phi_s $ according to Formula~(\ref{eq:sel}) \;
\end{algorithm}

\subsubsection{\textbf{Template Screening}} 

Considering that the number of potential templates (i.e. $ 2^L $) is still very large, we expect to further reduce the computational complexity in searching $\phi^*$.
To this end, we design an Exposure Template Search~(ETS) algorithm to effectively screen out a set of sub-optimal candidate templates, as illustrated in Algorithm~\ref{ag:beam}.
The basic idea is based on the Beam Search Algorithm~\cite{steinbiss1994improvements} and we use a tree model to represent the template search process.
As illustrated in Figure~\ref{fig:beam}, each tree node represents a sub-template (denoted by $\kappa$) whose length is equal to the depth of the corresponding layer (Line 4 to Line 11 in Algorithm~\ref{ag:beam}).
With the growth of the tree, we will iteratively prune some nodes to control the tree size (Line 12 to Line 14 in Algorithm~\ref{ag:beam}).
The pruning rule consists of the following two parts:
\begin{itemize}[leftmargin=*]
	\item \textbf{Eliminating unqualified nodes}: removing the nodes violating the request-level constraints;
	\item \textbf{Retaining top-$B$ nodes}: ranking the nodes by descending $ \Delta v $ and only retaining the top-$ B $ nodes.
\end{itemize}
At each layer, the number of remaining nodes is controlled by a beam size $ B $, i.e., the top-$B$ nodes are reserved until the tree depth reaches the request length $ L $.
The complexity of exposure template search is $ \mathcal{O}(L\times B) $, which is further significantly reduced compared with the previous $ \mathcal{O}(2^L) $.  

Within the final set of the well-chosen templates, denoted by $ \Phi_L $, the optimal exposure template $ \phi_s^* $ for $ s $ is determined according to the formula~(\ref{eq:kvim}), i.e., $\phi_s^* = \arg\max_{\phi \in \Phi_L } \Delta v(s|\phi) $.
Then the final exposure template
\begin{align} \label{eq:sel}
	\phi_s =
	\begin{cases}
		\phi_s^*, \text{ \ \ if } \rho(s|\phi_s^*) > \rho_{{}_{\rm THRES}}, \\
		\phi^0, \text{\ \ otherwise},
	\end{cases}
\end{align}
where $ \rho(s|\phi) = \frac{v(s|\phi)}{w(s|\phi)} $ according to (\ref{eq:vwr}).
The ETS algorithm can flexibly balance the optimality~(by selecting top-$ B $ nodes) and computational complexity~(by adjusting the beam size $ B $).

\subsection{Application-level Optimization with Real-time Feedback Control}\label{sec:pid}

However, the estimated $\rho_{{}_{\rm THRES}}$ might be different from the optimal $ \rho^*_{{}_{\rm THRES}} $.
When $ \rho_{{}_{\rm THRES}}  <  \rho^*_{{}_{\rm THRES}}  $, superfluous undesired ads are exposed and thus violate the application-level constraint (i.e., $ m > m^* $).
When $ \rho_{{}_{\rm THRES}}  > \rho^*_{{}_{\rm THRES}}  $, it leads to the loss of application revenue due to a reduced supply of ads exposures (i.e., $ m < m^* $).
Thus, we should dynamically adjust $ \rho_{{}_{\rm THRES}} $ to make actual monetization rate $ m $ close to the target $ m^* $.

We introduce a feedback control method~\cite{hagglund1995pid}  to timely adjust $ \rho_{{}_{\rm THRES}}$, keeping it close to the optimal $\rho^*_{{}_{\rm THRES}}$ and adapting to the fluctuating online environment.
In detail, we assume that $ \Delta t $ is a time interval to update $ \rho_{{}_{\rm THRES}} $.
At time step $ t $, we will calculate the monetization rate~(denoted by $ m^{(t-\Delta t : t)} $) over the requests within $ t-\Delta t $ to $ t $. 
Accordingly, $ \rho_{{}_{\rm THRES}} $ is updated as follows:
\begin{align}\label{eq:pid}
	\rho^{(t)}_{{}_{\rm THRES}} \longleftarrow \rho^{(t-\Delta t)}_{{}_{\rm THRES}}  \times \left[ 1 + \gamma \times \left(  \frac{m^{(t-\Delta t:t)}}{m^*} - 1 \right)  \right],
\end{align}
where $ \gamma $ is the learning rate.
Thus, when the actual $ m $ exceeds~(or is less than) $ m^* $, $ \rho_{{}_{\rm THRES}} $ will be increased~(or decreased) towards $ \rho^*_{{}_{\rm THRES}} $ to reduce the difference between $ m$ and $ m^* $.
In online experiments, we demonstrate that the real-time $ \rho_{{}_{\rm THRES}} $-adjustment greatly enhances the stability of monetization rate $ m $.

\subsection{Online Deployment}

\begin{figure}[!tp] 
	\centering
	\includegraphics[width=\linewidth]{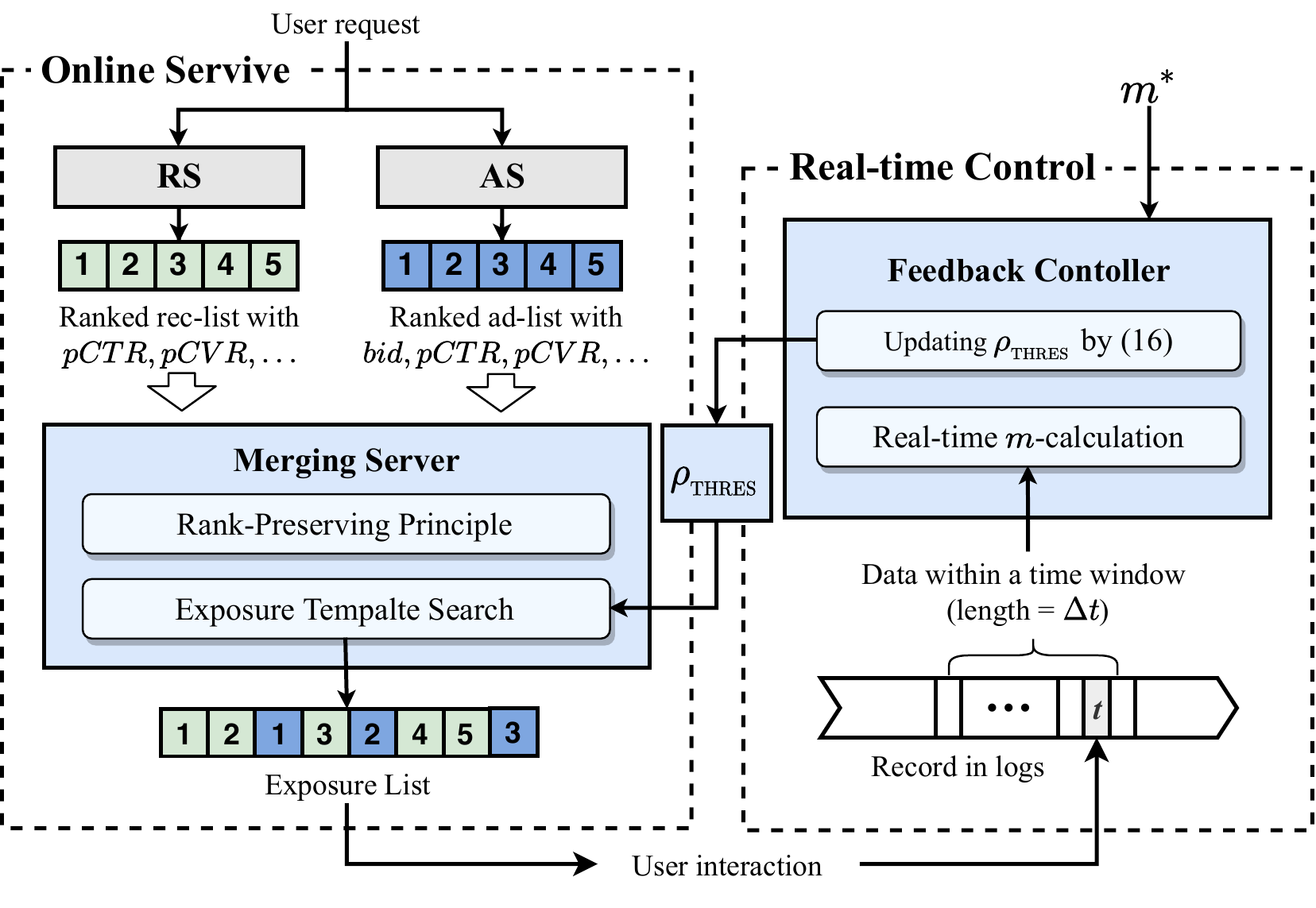}
	\caption{The pipeline of online service for HCA2E. }
	\label{fig:kvim-sys}
\end{figure}

HCA2E can reliably be deployed for online services, as shown in Figure \ref{fig:kvim-sys}.
The key ingredients are described as follows:
\begin{itemize}[leftmargin=*]
	\item
	In the online exposure service, once a user request arrives, RS and AS respectively process it as described in Section \ref{sec:back}.
	Then MS merges the two ranked lists~(with predicted indicators) according to Algorithm \ref{ag:beam}.
	Exposure results will be recorded in logs.
	\item
	Every period of $\Delta t$, the real-time controller will calculate the latest monetization rate $ m^{(t-\Delta t : t)} $ and adjust $ \rho_{{}_{\rm THRES}} $ according to Formula~(\ref{eq:pid}).
\end{itemize}
The initial threshold of VPW could be estimated by offline simulation with historical log data.
HCA2E has been deployed on a leading e-commerce application Taobao to serve users daily.

\section{Experiments}\label{sec:exp}
In this work, we conduct offline and online experiments based on the feed platform of Taobao.
We also claim that HCA2E can be applied in other types of feed applications.

\subsection{Experiment Settings}

\subsubsection{\textbf{Simulation Settings}}
In offline experiments, we set up a feed simulator that replays real log data to simulate users' activities and application responses.
Each data point corresponds to a user request and contains the information on recommendation candidates and ad candidates from RS and AS. 
The information mainly includes predicted indicators and homologous ranking scores.
Based on the candidate lists and input information, MS generates a blended list following the embedded exposure strategy. 
Further, we leverage a temporal data buffer to store requests within the updating interval of $\rho_{{}_{\rm THRES}}$ (i.e. $ \Delta t $). 
Every period of $ \Delta t $, $\rho_{{}_{\rm THRES}}$ is updated according to~(\ref{eq:pvr}) and the data buffer is initialized.
In the implementation, we collected about 10 million requests totally and set the updating time window with $ \Delta t = 10k $.
For each request, we set the number of exposure slots as 50, i.e. $ L = 50 $.
The top ad slot and minimum ad gap are set as 5 and 4 respectively.

\subsubsection{\textbf{Baseline Methods}}

We compare HCA2E with three baseline methods applied in industries.
To make fair comparisons, we will guarantee the baselines to satisfy the same constraints of monetization rate, top ad slot, and minimum ad gap as the HCA2E.
The baselines are briefly described as follows:
\begin{itemize}[leftmargin=*]
	\item  
	\textbf{Fixed} represents the fixed-position strategy, where the positions of recommended items and ads are manually pre-determined for every request. 
	The positions will be designed under the aforementioned different constraints.
	\item  
	\textbf{$ \beta $-WPO} is based on the Whole-Page Optimization~(WPO)~\cite{zhang2018whole}.
	WPO ranks recommended and ad candidates jointly according to the predefined ranking scores.
	To satisfy the $ m^* $-constraint, we introduce an adjustable variable $ \beta $ to control the ad proportion.
	\item 
	\textbf{$ \beta $-GEA} is based on the Gap Effect Algorithm (GEA)~\cite{yan2020ads}, which also employs a joint ranking score. 
	In addition, it takes the impact of adjacent ads' gap into account.
	Similar to $ \beta $-WPO, an adjustable variable $ \beta $ is introduced to control the ad proportion under $ m^* $.
\end{itemize}

\subsubsection{\textbf{Performance Metrics}} \label{sec:apd-metric}
We mainly focus on the following key performance indicators in the e-commerce feed:
\begin{itemize}[leftmargin=*]
	\item  
	\textbf{Revenue~(REV)} is the revenue produced by all exposed ads.
	\item  
	\textbf{Gross Merchandise Volume~(GMV)} is the merchandise volume of all purchased items~(both recommendations and ads).
	\item  
	\textbf{Click~(CLK)} is the total number of clicked items by users.
	\item  
	\textbf{Click-Through Rate~(CTR)} is the ratio of clicks to exposures.
\end{itemize}

\subsection{Offline Experiments}

\begin{table*}[!tp]
	\caption{The offline performance~($ \alpha = 0.5 $).}
	\label{tab:perf-off}
	
	\renewcommand\arraystretch{1.2}
	\setlength{\tabcolsep}{2.5mm}{
		\begin{tabular}{c|ccc|ccc|ccc}
			\toprule
			\hline
			
			& \multicolumn{3}{c|}{$m^*$ = 8\%} 
			& \multicolumn{3}{c|}{$m^*$ = 10\%} 
			& \multicolumn{3}{c}{$m^*$ = 12\%}  \\
			\hline
			Method & $ m $          & $ \Delta REV\% $            & $ \Delta GMV\% $    
			& $ m $           & $ \Delta REV\% $            & $ \Delta GMV\% $ 
			& $ m $           & $ \Delta REV\% $             & $ \Delta GMV\% $ \\  
			\hline
			Fixed     &   8.23\% &        $ -- $         &  $ -- $             
			&    10.36\%               &       $ -- $        &    $ -- $               
			&     12.29\%             &          $ -- $         &   $ -- $   \\
			$ \beta $-WPO       
			&  7.84\%  &     6.82\%     &     2.18\%     
			&   10.15\%         &     4.62\%     &    1.92\%
			&    12.19\% 			&		3.41\%			& 1.69\%  \\
			$ \beta $-GEA 
			&  8.13\%    & 		8.57\%		  &      2.47\%
			&  10.18\%            &      6.58\%          &   2.21\%
			&   11.78\%            &      4.62\%           &    1.82\%     \\
			\hline
			HCA2E(B=1)          
			& 7.97\%   &     {13.58\%}               &    {2.68\%}
			&  10.01\%     & {10.45\%}    &   {2.64\%}                 
			&  12.01\%    &  { 7.78\%  } &   { 2.03\% }  \\
			
			HCA2E(B=3)    
			& 8.03\%   &     {15.18\%}               &    {2.67\%}
			&  9.96\%     & {12.32\%}    &   {2.70\%}                 
			&  12.04\%    &  { 8.78\%  } &   { 2.07\% }  \\
			
			HCA2E(B=5)          
			& 8.01\%   &     {18.04\%}               &    {2.78\%}
			&  9.99\%     & {13.42\%}    &   {2.78\%}                 
			&  11.96\%    &  { 9.61\%  } &  \textbf{ 2.10\% }  \\
			
			HCA2E(B=7)          
			& 7.99\%   &     \textbf{18.45\%}               &    \textbf{2.79\%}
			&  10.02\%     & \textbf{13.68\%}    &   \textbf{2.81\%}                 
			&  12.03\%    &  \textbf{ 9.81\%  } &   { 2.09\% }  \\

			\hline
			\bottomrule
		\end{tabular}
	}
\end{table*}

\begin{figure*}[tp] 
	\centering
	\includegraphics[width=0.95\linewidth]{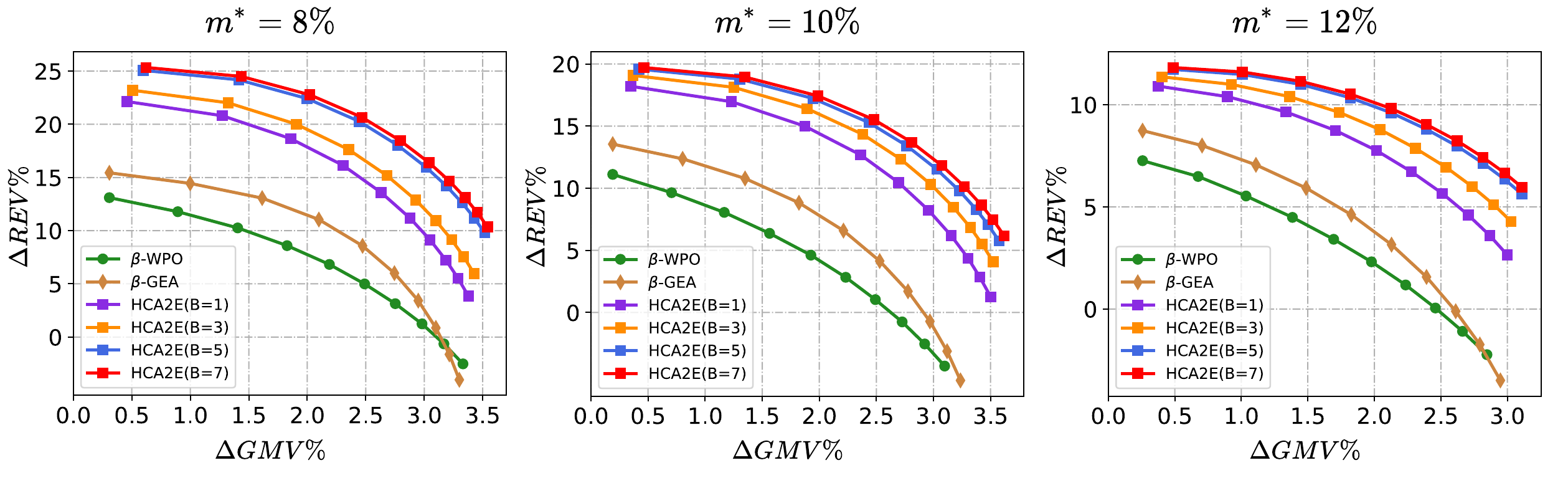}
	\caption{The Pareto curves under different $ m^* $.}
	\label{fig:pareto}
\end{figure*}

Within a specific simulation period, we observe the accumulative performances of the HCA2E and the three baseline methods.

\subsubsection{\textbf{Performance Evaluation}}

Under different target values of monetization rate ~($ m^* =$ 8\%, 10\%, 12\%), we evaluate the offline performance with REV and GMV jointly, respectively as the advertising-side and recommendation-side key indicators.
In this work, we measure the performance indicators with an advantage formula, calculated by the relative metric increment between an adaptive strategy and Fixed.
For example, the advantage of REV, denoted by $ \Delta REV \% $, is calculated as
\begin{align}\label{eq:drev}
	\Delta REV \%  = \frac{REV- REV_{\rm Fixed}}{REV_{\rm Fixed}} \times 100\%.
\end{align}

Given a trade-off parameter $ \alpha $, we can obtain a corresponding tuple of $ \Delta REV \%$ and $ \Delta GMV \% $.
Table  \ref{tab:perf-off} records the performances of different algorithms with $ \alpha = 0.5 $.
For adaptive exposure strategies, varying the value of $ \alpha $ can produce different performance tuples~($ \Delta REV \%$,  $ \Delta GMV \% $), known as Pareto-optimal solutions~\cite{lin2019pareto}.
We select several values of $ \alpha $  to obtain a set of solutions and draw the Pareto-optimal trade-off curve as shown in Figure \ref{fig:pareto}, where
$ \alpha $ is varied within $ \left[0.1, 0.2, 0.3, \cdots, 1.0 \right] $,
and $ \Delta REV\% $/$ \Delta GMV \% $ is corresponding to y-axis/x-axis.
From Table  \ref{tab:perf-off}  and Figure \ref{fig:pareto}, we can make the following observations:

\textbf{1)} 
As $ \alpha $ varies, HCA2E algorithms with four values of beam size (i.e., $ B = 1,3,5,7 $) could have better Pareto-optimal curves, i.e., when reaching the same GMV~(or REV), HCA2E can achieve higher REV~(or GMV) than $ \beta $-WPO and $ \beta $-GEA.
The Pareto-optimal curves verify that HCA2E achieves the overall performance improvement compared with the baseline methods due to the optimization towards the application-level performance.

\textbf{\textbf{2)}} 
Within a certain range, the increase of beam size $ B $ could boost the performance of HCA2E.
A larger $ B $ could provide more possible candidate templates, and thereby some better solutions are taken into account.
However, such improvement by increasing $ B $ is limited.
For example, HCA2E($ B = 5 $) and HCA2E($ B = 7 $) achieve closed performances corresponding to the almost overlapping curves.
This is because the request value is mostly determined by several top slots due to higher exposure probability.
Thus, the ETS algorithm does not need a large $ B $ to restore the sub-templates.
In this view, we can adjust $ B $ to achieve a good balance between the performance and the search complexity.

\textbf{3)}
We also find that the real-time feedback control method can effectively stabilize the actual monetization rate.
When the given $ m^* = $ 8\%, 10\% or 12\%, actual $ m $ of HCA2E could have smaller bias to $ m^* $ than baselines.

\subsubsection{\textbf{Ad-exposure Distribution Analysis}}

\begin{figure}[t] 
	\centering
	\includegraphics[width=0.95\linewidth]{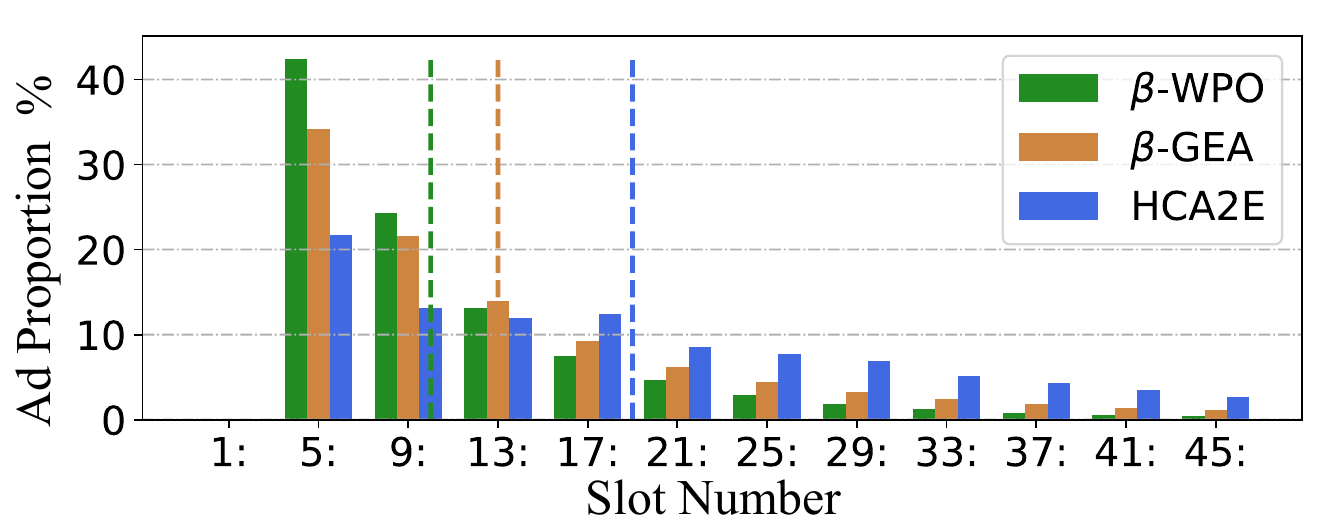}
	\caption{The ad-exposure distribution. The height of each bar on "$ k: $" indicates the ad proportion within slots $ k $ to $ k+3 $.
		There is no ad at the first four slots due to the top ad slot constraint.
		The dotted lines indicate average ad positions.}
	\label{fig:slots}
\end{figure}

Generally, ads at top slots will be viewed/clicked more probably and thus obtain more revenue.
It is usually desirable to achieve higher revenue but expose ads at lower slots due to the cost of GMV.
Thus, we analyzed the ad-exposure distribution, measured by the percentage of ad exposures on each slot within total ad exposures, to account for whether the performance improvements of HCA2E owe to that more ads are placed at front positions.

We select that $ m^* = 10\% $ and $ \alpha = 0.5 $.
For $ \beta $-WPO, $ \beta $-GEA, and HCA2E($ B = 5 $), the ad-exposure distributions are shown in Figure \ref{fig:slots}.
We calculate the percentage of ad exposures for every 4~(equal to the minimum ad gap) slots.
The result shows that HCA2E has a more even ad exposure distribution than the others.
Within top-12 slots, HCA2E exposes much fewer ads than $ \beta $-WPO and $ \beta $-GEA. 
Below the 29-th slot, HCA2E still exposes a certain number of ads, but there are nearly none for $ \beta $-WPO or $ \beta $-GEA.
Further, we calculate the average ad positions for each approach (corresponding to the dotted lines drawn in Figure \ref{fig:slots}).
We can find that, despite achieving higher REV, the average ad position of HCA2E is lower than the others.
Thus, we demonstrate that the higher REV of HCA2E does not rely on placing more ads at top slots, and HCA2E indeed achieves higher efficiency for ad exposure.

\subsection{Online A/B Testing}
In online experiments, we successfully deploy HCA2E on a feed product of Taobao, named \textit{Guess What You Like}.
Here we select the results from six major scenarios, located on various pages of the Taobao application, such as Homepage, Payment page, Cart page,  etc.
We observe the accumulative performance of HCA2E over two weeks and compare it with the Fixed's performance through online A/B testing.

\begin{table}[!tp] 
	\caption{The online performance in six feed scenarios of Taobao. The arrow "$ \rightarrow $" is from Fixed to HCA2E.} 
	\label{tab:perf-on}
	
	\renewcommand\arraystretch{1.2}
	\setlength{\tabcolsep}{1.2mm}{
		\begin{tabular}{c|cc|cc|c}
			\toprule
			\hline
			
			Scenario & $ \Delta REV\% $  & $ \Delta CTR^{\rm ad}\%$  & $ \Delta GMV\%$  & $ \Delta CLK\% $  &  $ \#Slot^{\rm ad}_{\rm avg} $ \\  
			\hline
			Homepage    &        9.24\%            & 6.24\%      &    1.09\%   &  0.22\%  &     $ 33 \rightarrow 120 $       \\
			Collection     &         6.20\%   & 4.63\%          &   3.89\%        &   1.25\%       &     $ 27 \rightarrow 76 $      \\
			Cart  			  &        6.57\% &6.19\%         &     4.96\%      &   0.46\%     &   $ 27 \rightarrow 71 $        \\
			Payment       &        5.91\%     & 4.35\%     &      1.01\%    &    1.03\%     &     $ 22 \rightarrow 47 $      \\
			Order List     &      11.64\%     &    12.33\%    & 8.65\%  &    1.01\%     
			&  	$ 26 \rightarrow 53 $         \\
			Logistics       &     3.58\%         &    1.79\%     &      5.88\%     &  0.62\%  &      $ 23 \rightarrow 53 $       \\
			
			\hline
			\bottomrule
		\end{tabular}
	}
\end{table}

\subsubsection{\textbf{Performance Evaluation}}

In response to the different characteristics of these scenarios, we will adjust the trade-off parameter $ \alpha $ of HCA2E to guarantee positive performance.
Here, mainly concerned indicators include REV, advertising CTR~($ CTR^{\rm ad} $), GMV, and CLK.
Similar to the metric~(\ref{eq:drev}), they are expressed as the indicator advantage to Fixed.
For all of these scenarios, the online results within a week, recorded in Table \ref{tab:perf-on}, also show the effectiveness and superiority of HCA2E compared with the Fixed.
Due to the differences among these scenarios~(e.g. traffic difference, user preference, and different target monetization rates), the performance indicators are improved to different degrees.
In addition, we compare the average ad positions~($ \#Slot^{\rm ad}_{\rm avg} $) of HCA2E and Fixed, as shown in the last column of Table \ref{tab:perf-on}.
We find that $ \#Slot^{\rm ad}_{\rm avg} $ of HCA2E is much lower than Fixed, further demonstrating higher ad exposure efficiency of HCA2E.

\begin{figure}[tp] 
	\centering
	\includegraphics[width=0.95\linewidth]{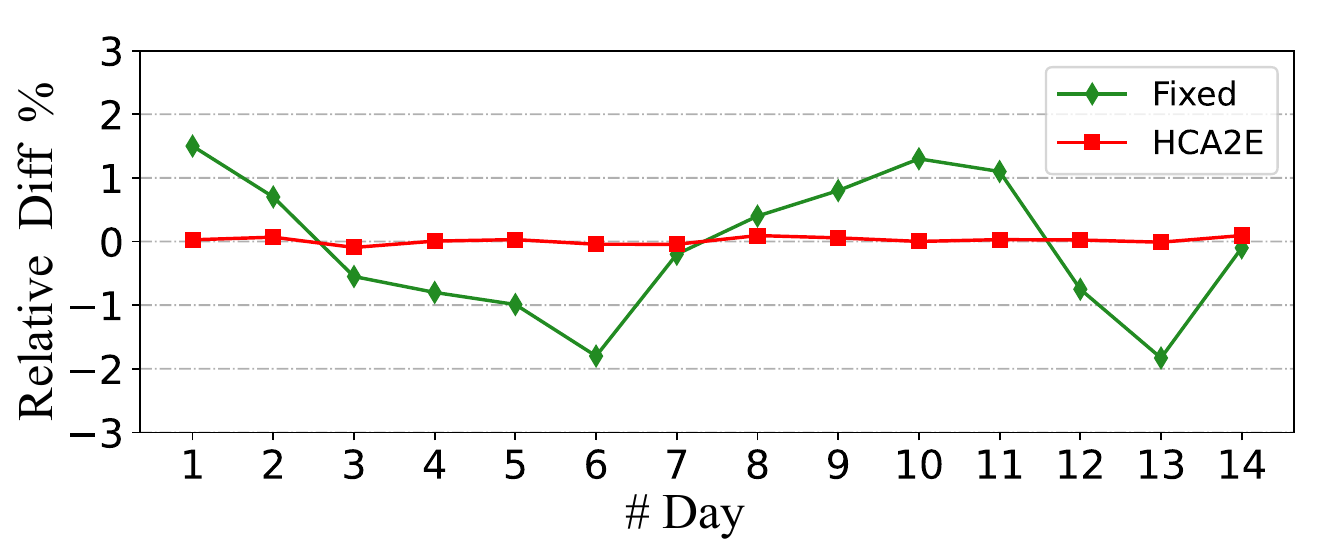}
	\caption{The fluctuations of $ m $ within two weeks. The y-axis represents the relative difference of $ m $ to $ m^* $. We can find that the stability of $ m $ is enhanced by introducing  the feedback control method.}
	\label{fig:pvr}
\end{figure}

\subsubsection{\textbf{Robustness Analysis of monetization rate.}}

Furthermore, for HCA2E and Fixed, we observe the fluctuations of the monetization rate in Homepage.
We aim to demonstrate the effectiveness of introducing the feedback control method to stabilize the monetization rate.
Figure \ref{fig:pvr} shows the monetization rate fluctuation results of HCA2E and Fixed within two weeks, where the y-axis corresponds to the relative difference between actual $ m $ and target $ m^* $.
We find that $ m $ of the Fixed strategy has a larger variation margin.
Though the ad positions are predetermined in Fixed, $ m $ within different periods could be varying due to the differences in users' exploring depth on different requests.
Within all 14 days, the curve of HCA2E is consistently close to the zero-line~(i.e., $ m $ is close to $ m^* $).
Thus, HCA2E enhances a accumulative stability of $ m $ via the real-time $ \rho_{{}_{\rm THRES}} $-control.  

\section{Related Works}

Earlier studies for adaptive ad exposure only focus on one of the following problems:
whether to expose ads~\cite{broder2008swing}, how many ads to expose~\cite{wang2011learning} and where to expose ads~\cite{zhang2018whole}.
Recently, adaptive ad exposure has been studied integrally.
Rule-based algorithms usually blend items according to the preset re-ranking rules.
For example, the Whole-Page Optimization~(WPO) method calculates a unified ranking score for both recommended and ad items, and then it ranks items in the order of descending ranking scores~\cite{zhang2018whole}. 
The Gap Effect Algorithm~(GEA)~\cite{yan2020ads} is proposed to maximize the ad revenue of each request under a user engagement constraint.
The GEA also re-ranks mixed items through mapping ad-utility into a comparable metric with rec-utility.
Different from WPO, GEA would consider the gap effect between consecutive ads.
More complex learning-based methods mostly attempt to formulate adaptive ad exposure as a Markov Decision Process~(MDP)~\cite{van2012reinforcement} and learn the strategy with effective reinforcement learning approaches~\cite{wang2019learning, zhao2020jointly, zhao2021dear, liao2021cross}.
In the MDP, a state is represented by contextual features of candidate items; 
an action corresponds to an exposure result; 
the reward function could be calculated by the feedback of a user request.
Such methods allow the strategy optimization under an end-to-end learning pattern,
where deep networks are usually utilized.
Despite enjoying good performances in offline simulations, learning-based methods 
usually suffer from 1) a large action space, 2) massive model parameters, and 3) inflexible objective transitions.
Thus, they are hardly deployed in large-scale applications.

\section{Conclusion}

In this paper, we formulate the adaptive ad exposure problem in feeds as a Dynamic Knapsack Problem~(DKP), towards application-level performance optimization under hierarchical constraints.
We propose a new approach, Hierarchically Constrained Adaptive Ad Exposure~(HCA2E), which possesses the desirable game-theoretical properties, computational efficiency, and performance robustness.
The offline and online evaluations demonstrate the performance superiority of HCA2E.
HCA2E has been deployed on Taobao to serve millions of users daily and achieved significant performance improvements.

\begin{acks}
This work was supported in part by Science and Technology Innovation 2030 –“New Generation Artificial Intelligence” Major Project No. 2018AAA0100905, in part by Alibaba Group through Alibaba Innovation Research Program, in part by China NSF grant No. 62132018, 61902248, 62025204, in part by Shanghai Science and Technology fund 20PJ1407900. The opinions, findings, conclusions, and recommendations expressed in this paper are those of the authors and do not necessarily reflect the views of the funding agencies or the government.
\end{acks}

\clearpage

\bibliographystyle{ACM-Reference-Format}
\balance
\bibliography{sample-base}


\begin{thebibliography}{28}


\ifx \showCODEN    \undefined \def \showCODEN     #1{\unskip}     \fi
\ifx \showDOI      \undefined \def \showDOI       #1{#1}\fi
\ifx \showISBNx    \undefined \def \showISBNx     #1{\unskip}     \fi
\ifx \showISBNxiii \undefined \def \showISBNxiii  #1{\unskip}     \fi
\ifx \showISSN     \undefined \def \showISSN      #1{\unskip}     \fi
\ifx \showLCCN     \undefined \def \showLCCN      #1{\unskip}     \fi
\ifx \shownote     \undefined \def \shownote      #1{#1}          \fi
\ifx \showarticletitle \undefined \def \showarticletitle #1{#1}   \fi
\ifx \showURL      \undefined \def \showURL       {\relax}        \fi
\providecommand\bibfield[2]{#2}
\providecommand\bibinfo[2]{#2}
\providecommand\natexlab[1]{#1}
\providecommand\showeprint[2][]{arXiv:#2}

\bibitem[\protect\citeauthoryear{Aggarwal, Muthukrishnan, P{\'a}l, and
  P{\'a}l}{Aggarwal et~al\mbox{.}}{2009}]%
        {aggarwal2009general}
\bibfield{author}{\bibinfo{person}{Gagan Aggarwal}, \bibinfo{person}{S
  Muthukrishnan}, \bibinfo{person}{D{\'a}vid P{\'a}l}, {and}
  \bibinfo{person}{Martin P{\'a}l}.} \bibinfo{year}{2009}\natexlab{}.
\newblock \showarticletitle{General auction mechanism for search advertising}.
  In \bibinfo{booktitle}{\emph{Proceedings of the 18th international conference
  on World wide web}}. \bibinfo{pages}{241--250}.
\newblock


\bibitem[\protect\citeauthoryear{Broder, Ciaramita, Fontoura, Gabrilovich,
  Josifovski, Metzler, Murdock, and Plachouras}{Broder et~al\mbox{.}}{2008}]%
        {broder2008swing}
\bibfield{author}{\bibinfo{person}{Andrei Broder},
  \bibinfo{person}{Massimiliano Ciaramita}, \bibinfo{person}{Marcus Fontoura},
  \bibinfo{person}{Evgeniy Gabrilovich}, \bibinfo{person}{Vanja Josifovski},
  \bibinfo{person}{Donald Metzler}, \bibinfo{person}{Vanessa Murdock}, {and}
  \bibinfo{person}{Vassilis Plachouras}.} \bibinfo{year}{2008}\natexlab{}.
\newblock \showarticletitle{To swing or not to swing: learning when (not) to
  advertise}. In \bibinfo{booktitle}{\emph{Proceedings of the 17th ACM
  conference on information and knowledge management}}.
  \bibinfo{pages}{1003--1012}.
\newblock


\bibitem[\protect\citeauthoryear{Chen, Jin, Zhang, Pan, Niu, Yu, Wang, Li, Xu,
  and Gai}{Chen et~al\mbox{.}}{2019}]%
        {chen2019learning}
\bibfield{author}{\bibinfo{person}{Dagui Chen}, \bibinfo{person}{Junqi Jin},
  \bibinfo{person}{Weinan Zhang}, \bibinfo{person}{Fei Pan},
  \bibinfo{person}{Lvyin Niu}, \bibinfo{person}{Chuan Yu}, \bibinfo{person}{Jun
  Wang}, \bibinfo{person}{Han Li}, \bibinfo{person}{Jian Xu}, {and}
  \bibinfo{person}{Kun Gai}.} \bibinfo{year}{2019}\natexlab{}.
\newblock \showarticletitle{Learning to Advertise for Organic Traffic
  Maximization in E-Commerce Product Feeds}. In
  \bibinfo{booktitle}{\emph{Proceedings of the 28th ACM International
  Conference on Information and Knowledge Management}}.
  \bibinfo{pages}{2527--2535}.
\newblock


\bibitem[\protect\citeauthoryear{Dantzig}{Dantzig}{1955}]%
        {dantzig1955discrete}
\bibfield{author}{\bibinfo{person}{GB Dantzig}.}
  \bibinfo{year}{1955}\natexlab{}.
\newblock \showarticletitle{Discrete variable extremum problems}. In
  \bibinfo{booktitle}{\emph{JOURNAL OF THE OPERATIONS RESEARCH SOCIETY OF
  AMERICA}}, Vol.~\bibinfo{volume}{3}. \bibinfo{pages}{560--560}.
\newblock


\bibitem[\protect\citeauthoryear{Dizdar, Gershkov, and Moldovanu}{Dizdar
  et~al\mbox{.}}{2011}]%
        {dizdar2011revenue}
\bibfield{author}{\bibinfo{person}{Deniz Dizdar}, \bibinfo{person}{Alex
  Gershkov}, {and} \bibinfo{person}{Benny Moldovanu}.}
  \bibinfo{year}{2011}\natexlab{}.
\newblock \showarticletitle{Revenue maximization in the dynamic knapsack
  problem}.
\newblock \bibinfo{journal}{\emph{Theoretical Economics}} \bibinfo{volume}{6},
  \bibinfo{number}{2} (\bibinfo{year}{2011}), \bibinfo{pages}{157--184}.
\newblock


\bibitem[\protect\citeauthoryear{Geyik, Faleev, Shen, O'Donnell, and
  Kolay}{Geyik et~al\mbox{.}}{2016}]%
        {geyik2016joint}
\bibfield{author}{\bibinfo{person}{Sahin~Cem Geyik}, \bibinfo{person}{Sergey
  Faleev}, \bibinfo{person}{Jianqiang Shen}, \bibinfo{person}{Sean O'Donnell},
  {and} \bibinfo{person}{Santanu Kolay}.} \bibinfo{year}{2016}\natexlab{}.
\newblock \showarticletitle{Joint optimization of multiple performance metrics
  in online video advertising}. In \bibinfo{booktitle}{\emph{Proceedings of the
  22nd ACM SIGKDD International Conference on Knowledge Discovery and Data
  Mining}}. \bibinfo{pages}{471--480}.
\newblock


\bibitem[\protect\citeauthoryear{Hagglund and Astrom}{Hagglund and
  Astrom}{1995}]%
        {hagglund1995pid}
\bibfield{author}{\bibinfo{person}{Tore Hagglund} {and} \bibinfo{person}{Karl~J
  Astrom}.} \bibinfo{year}{1995}\natexlab{}.
\newblock \showarticletitle{PID controllers: theory, design, and tuning}.
\newblock \bibinfo{journal}{\emph{ISA-The Instrumentation, Systems, and
  Automation Society}} (\bibinfo{year}{1995}).
\newblock


\bibitem[\protect\citeauthoryear{Hao, Peng, Ma, Wang, Jin, Hao, Chen, Bai, Xie,
  Xu, et~al\mbox{.}}{Hao et~al\mbox{.}}{2020}]%
        {hao2020dynamic}
\bibfield{author}{\bibinfo{person}{Xiaotian Hao}, \bibinfo{person}{Zhaoqing
  Peng}, \bibinfo{person}{Yi Ma}, \bibinfo{person}{Guan Wang},
  \bibinfo{person}{Junqi Jin}, \bibinfo{person}{Jianye Hao},
  \bibinfo{person}{Shan Chen}, \bibinfo{person}{Rongquan Bai},
  \bibinfo{person}{Mingzhou Xie}, \bibinfo{person}{Miao Xu}, {et~al\mbox{.}}}
  \bibinfo{year}{2020}\natexlab{}.
\newblock \showarticletitle{Dynamic knapsack optimization towards efficient
  multi-channel sequential advertising}. In
  \bibinfo{booktitle}{\emph{International Conference on Machine Learning}}.
  PMLR, \bibinfo{pages}{4060--4070}.
\newblock


\bibitem[\protect\citeauthoryear{Jin, Song, Li, Gai, Wang, and Zhang}{Jin
  et~al\mbox{.}}{2018}]%
        {jin2018real}
\bibfield{author}{\bibinfo{person}{Junqi Jin}, \bibinfo{person}{Chengru Song},
  \bibinfo{person}{Han Li}, \bibinfo{person}{Kun Gai}, \bibinfo{person}{Jun
  Wang}, {and} \bibinfo{person}{Weinan Zhang}.}
  \bibinfo{year}{2018}\natexlab{}.
\newblock \showarticletitle{Real-time bidding with multi-agent reinforcement
  learning in display advertising}. In \bibinfo{booktitle}{\emph{Proceedings of
  the 27th ACM International Conference on Information and Knowledge
  Management}}. \bibinfo{pages}{2193--2201}.
\newblock


\bibitem[\protect\citeauthoryear{Liao, Wang, Wu, Shi, Zhang, Wang, Wang, and
  Wang}{Liao et~al\mbox{.}}{2021}]%
        {liao2021cross}
\bibfield{author}{\bibinfo{person}{Guogang Liao}, \bibinfo{person}{Ze Wang},
  \bibinfo{person}{Xiaoxu Wu}, \bibinfo{person}{Xiaowen Shi},
  \bibinfo{person}{Chuheng Zhang}, \bibinfo{person}{Yongkang Wang},
  \bibinfo{person}{Xingxing Wang}, {and} \bibinfo{person}{Dong Wang}.}
  \bibinfo{year}{2021}\natexlab{}.
\newblock \showarticletitle{Cross DQN: Cross Deep Q Network for Ads Allocation
  in Feed}.
\newblock \bibinfo{journal}{\emph{arXiv preprint arXiv:2109.04353}}
  (\bibinfo{year}{2021}).
\newblock


\bibitem[\protect\citeauthoryear{Lin, Zhen, Li, Zhang, and Kwong}{Lin
  et~al\mbox{.}}{2019}]%
        {lin2019pareto}
\bibfield{author}{\bibinfo{person}{Xi Lin}, \bibinfo{person}{Hui-Ling Zhen},
  \bibinfo{person}{Zhenhua Li}, \bibinfo{person}{Qing-Fu Zhang}, {and}
  \bibinfo{person}{Sam Kwong}.} \bibinfo{year}{2019}\natexlab{}.
\newblock \showarticletitle{Pareto multi-task learning}.
\newblock \bibinfo{journal}{\emph{Advances in neural information processing
  systems}}  \bibinfo{volume}{32} (\bibinfo{year}{2019}),
  \bibinfo{pages}{12060--12070}.
\newblock


\bibitem[\protect\citeauthoryear{Liu, Yu, Zhang, Zheng, Rong, Lv, Huo, Wang,
  Chen, Xu, Wu, Chen, and Zhu}{Liu et~al\mbox{.}}{2021}]%
        {liu2021neural}
\bibfield{author}{\bibinfo{person}{Xiangyu Liu}, \bibinfo{person}{Chuan Yu},
  \bibinfo{person}{Zhilin Zhang}, \bibinfo{person}{Zhenzhe Zheng},
  \bibinfo{person}{Yu Rong}, \bibinfo{person}{Hongtao Lv}, \bibinfo{person}{Da
  Huo}, \bibinfo{person}{Yiqing Wang}, \bibinfo{person}{Dagui Chen},
  \bibinfo{person}{Jian Xu}, \bibinfo{person}{Fan Wu}, \bibinfo{person}{Guihai
  Chen}, {and} \bibinfo{person}{Xiaoqiang Zhu}.}
  \bibinfo{year}{2021}\natexlab{}.
\newblock \showarticletitle{Neural Auction: End-to-End Learning of Auction
  Mechanisms for E-Commerce Advertising}. In \bibinfo{booktitle}{\emph{{KDD}
  '21, Singapore}}. \bibinfo{publisher}{{ACM}}, \bibinfo{pages}{3354--3364}.
\newblock


\bibitem[\protect\citeauthoryear{Myerson}{Myerson}{1981}]%
        {myerson1981optimal}
\bibfield{author}{\bibinfo{person}{Roger~B Myerson}.}
  \bibinfo{year}{1981}\natexlab{}.
\newblock \showarticletitle{Optimal auction design}.
\newblock \bibinfo{journal}{\emph{Mathematics of operations research}}
  \bibinfo{volume}{6}, \bibinfo{number}{1} (\bibinfo{year}{1981}),
  \bibinfo{pages}{58--73}.
\newblock


\bibitem[\protect\citeauthoryear{Nazerzadeh, Saberi, and Vohra}{Nazerzadeh
  et~al\mbox{.}}{2013}]%
        {nazerzadeh2013dynamic}
\bibfield{author}{\bibinfo{person}{Hamid Nazerzadeh}, \bibinfo{person}{Amin
  Saberi}, {and} \bibinfo{person}{Rakesh Vohra}.}
  \bibinfo{year}{2013}\natexlab{}.
\newblock \showarticletitle{Dynamic pay-per-action mechanisms and applications
  to online advertising}.
\newblock \bibinfo{journal}{\emph{Operations Research}} \bibinfo{volume}{61},
  \bibinfo{number}{1} (\bibinfo{year}{2013}), \bibinfo{pages}{98--111}.
\newblock


\bibitem[\protect\citeauthoryear{Salkin and De~Kluyver}{Salkin and
  De~Kluyver}{1975}]%
        {salkin1975knapsack}
\bibfield{author}{\bibinfo{person}{Harvey~M Salkin} {and}
  \bibinfo{person}{Cornelis~A De~Kluyver}.} \bibinfo{year}{1975}\natexlab{}.
\newblock \showarticletitle{The knapsack problem: a survey}.
\newblock \bibinfo{journal}{\emph{Naval Research Logistics Quarterly}}
  \bibinfo{volume}{22}, \bibinfo{number}{1} (\bibinfo{year}{1975}),
  \bibinfo{pages}{127--144}.
\newblock


\bibitem[\protect\citeauthoryear{Steinbiss, Tran, and Ney}{Steinbiss
  et~al\mbox{.}}{1994}]%
        {steinbiss1994improvements}
\bibfield{author}{\bibinfo{person}{Volker Steinbiss},
  \bibinfo{person}{Bach-Hiep Tran}, {and} \bibinfo{person}{Hermann Ney}.}
  \bibinfo{year}{1994}\natexlab{}.
\newblock \showarticletitle{Improvements in beam search}. In
  \bibinfo{booktitle}{\emph{Third international conference on spoken language
  processing}}.
\newblock


\bibitem[\protect\citeauthoryear{Sutton and Barto}{Sutton and Barto}{2018}]%
        {sutton2018reinforcement}
\bibfield{author}{\bibinfo{person}{Richard~S Sutton} {and}
  \bibinfo{person}{Andrew~G Barto}.} \bibinfo{year}{2018}\natexlab{}.
\newblock \bibinfo{booktitle}{\emph{Reinforcement learning: An introduction}}.
\newblock \bibinfo{publisher}{MIT press}.
\newblock


\bibitem[\protect\citeauthoryear{Van~Otterlo and Wiering}{Van~Otterlo and
  Wiering}{2012}]%
        {van2012reinforcement}
\bibfield{author}{\bibinfo{person}{Martijn Van~Otterlo} {and}
  \bibinfo{person}{Marco Wiering}.} \bibinfo{year}{2012}\natexlab{}.
\newblock \showarticletitle{Reinforcement learning and markov decision
  processes}.
\newblock In \bibinfo{booktitle}{\emph{Reinforcement learning}}.
  \bibinfo{publisher}{Springer}, \bibinfo{pages}{3--42}.
\newblock


\bibitem[\protect\citeauthoryear{Vickrey}{Vickrey}{1961}]%
        {vickrey1961counterspeculation}
\bibfield{author}{\bibinfo{person}{William Vickrey}.}
  \bibinfo{year}{1961}\natexlab{}.
\newblock \showarticletitle{Counterspeculation, auctions, and competitive
  sealed tenders}.
\newblock \bibinfo{journal}{\emph{The Journal of finance}}
  \bibinfo{volume}{16}, \bibinfo{number}{1} (\bibinfo{year}{1961}),
  \bibinfo{pages}{8--37}.
\newblock


\bibitem[\protect\citeauthoryear{Wang, Li, Tang, Zhang, Chen, and Ru}{Wang
  et~al\mbox{.}}{2011}]%
        {wang2011learning}
\bibfield{author}{\bibinfo{person}{Bo Wang}, \bibinfo{person}{Zhaonan Li},
  \bibinfo{person}{Jie Tang}, \bibinfo{person}{Kuo Zhang},
  \bibinfo{person}{Songcan Chen}, {and} \bibinfo{person}{Liyun Ru}.}
  \bibinfo{year}{2011}\natexlab{}.
\newblock \showarticletitle{Learning to advertise: how many ads are enough?}.
  In \bibinfo{booktitle}{\emph{Pacific-Asia Conference on Knowledge Discovery
  and Data Mining}}. Springer, \bibinfo{pages}{506--518}.
\newblock


\bibitem[\protect\citeauthoryear{Wang, Jin, Hao, Chen, Yu, Zhang, Wang, Hao,
  Wang, Li, et~al\mbox{.}}{Wang et~al\mbox{.}}{2019}]%
        {wang2019learning}
\bibfield{author}{\bibinfo{person}{Weixun Wang}, \bibinfo{person}{Junqi Jin},
  \bibinfo{person}{Jianye Hao}, \bibinfo{person}{Chunjie Chen},
  \bibinfo{person}{Chuan Yu}, \bibinfo{person}{Weinan Zhang},
  \bibinfo{person}{Jun Wang}, \bibinfo{person}{Xiaotian Hao},
  \bibinfo{person}{Yixi Wang}, \bibinfo{person}{Han Li}, {et~al\mbox{.}}}
  \bibinfo{year}{2019}\natexlab{}.
\newblock \showarticletitle{Learning Adaptive Display Exposure for Real-Time
  Advertising}. In \bibinfo{booktitle}{\emph{Proceedings of the 28th ACM
  International Conference on Information and Knowledge Management}}.
  \bibinfo{pages}{2595--2603}.
\newblock


\bibitem[\protect\citeauthoryear{Wilkens, Cavallo, and Niazadeh}{Wilkens
  et~al\mbox{.}}{2017}]%
        {wilkens2017gsp}
\bibfield{author}{\bibinfo{person}{Christopher~A Wilkens},
  \bibinfo{person}{Ruggiero Cavallo}, {and} \bibinfo{person}{Rad Niazadeh}.}
  \bibinfo{year}{2017}\natexlab{}.
\newblock \showarticletitle{GSP: the cinderella of mechanism design}. In
  \bibinfo{booktitle}{\emph{Proceedings of the 26th International Conference on
  World Wide Web}}. \bibinfo{pages}{25--32}.
\newblock


\bibitem[\protect\citeauthoryear{Yan, Xu, Tiwana, and Chatterjee}{Yan
  et~al\mbox{.}}{2020}]%
        {yan2020ads}
\bibfield{author}{\bibinfo{person}{Jinyun Yan}, \bibinfo{person}{Zhiyuan Xu},
  \bibinfo{person}{Birjodh Tiwana}, {and} \bibinfo{person}{Shaunak
  Chatterjee}.} \bibinfo{year}{2020}\natexlab{}.
\newblock \showarticletitle{Ads Allocation in Feed via Constrained
  Optimization}. In \bibinfo{booktitle}{\emph{Proceedings of the 26th ACM
  SIGKDD International Conference on Knowledge Discovery \& Data Mining}}.
  \bibinfo{pages}{3386--3394}.
\newblock


\bibitem[\protect\citeauthoryear{Zhang, Wei, Meng, Hu, and Wang}{Zhang
  et~al\mbox{.}}{2018}]%
        {zhang2018whole}
\bibfield{author}{\bibinfo{person}{Weiru Zhang}, \bibinfo{person}{Chao Wei},
  \bibinfo{person}{Xiaonan Meng}, \bibinfo{person}{Yi Hu}, {and}
  \bibinfo{person}{Hao Wang}.} \bibinfo{year}{2018}\natexlab{}.
\newblock \showarticletitle{The whole-page optimization via dynamic ad
  allocation}. In \bibinfo{booktitle}{\emph{Companion Proceedings of the The
  Web Conference 2018}}. \bibinfo{pages}{1407--1411}.
\newblock


\bibitem[\protect\citeauthoryear{Zhang, Liu, Zheng, Zhang, Xu, Pan, Yu, Wu, Xu,
  and Gai}{Zhang et~al\mbox{.}}{2021}]%
        {zhang2021optimizing}
\bibfield{author}{\bibinfo{person}{Zhilin Zhang}, \bibinfo{person}{Xiangyu
  Liu}, \bibinfo{person}{Zhenzhe Zheng}, \bibinfo{person}{Chenrui Zhang},
  \bibinfo{person}{Miao Xu}, \bibinfo{person}{Junwei Pan},
  \bibinfo{person}{Chuan Yu}, \bibinfo{person}{Fan Wu}, \bibinfo{person}{Jian
  Xu}, {and} \bibinfo{person}{Kun Gai}.} \bibinfo{year}{2021}\natexlab{}.
\newblock \showarticletitle{Optimizing Multiple Performance Metrics with Deep
  GSP Auctions for E-commerce Advertising}. In
  \bibinfo{booktitle}{\emph{Proceedings of the 14th ACM International
  Conference on Web Search and Data Mining}}. \bibinfo{pages}{993--1001}.
\newblock


\bibitem[\protect\citeauthoryear{Zhao, Gu, Zhang, Yang, Liu, Liu, and
  Tang}{Zhao et~al\mbox{.}}{2021}]%
        {zhao2021dear}
\bibfield{author}{\bibinfo{person}{Xiangyu Zhao}, \bibinfo{person}{Changsheng
  Gu}, \bibinfo{person}{Haoshenglun Zhang}, \bibinfo{person}{Xiwang Yang},
  \bibinfo{person}{Xiaobing Liu}, \bibinfo{person}{Hui Liu}, {and}
  \bibinfo{person}{Jiliang Tang}.} \bibinfo{year}{2021}\natexlab{}.
\newblock \showarticletitle{DEAR: Deep Reinforcement Learning for Online
  Advertising Impression in Recommender Systems}. In
  \bibinfo{booktitle}{\emph{Proceedings of the AAAI Conference on Artificial
  Intelligence}}, Vol.~\bibinfo{volume}{35}. \bibinfo{pages}{750--758}.
\newblock


\bibitem[\protect\citeauthoryear{Zhao, Zheng, Yang, Liu, and Tang}{Zhao
  et~al\mbox{.}}{2020}]%
        {zhao2020jointly}
\bibfield{author}{\bibinfo{person}{Xiangyu Zhao}, \bibinfo{person}{Xudong
  Zheng}, \bibinfo{person}{Xiwang Yang}, \bibinfo{person}{Xiaobing Liu}, {and}
  \bibinfo{person}{Jiliang Tang}.} \bibinfo{year}{2020}\natexlab{}.
\newblock \showarticletitle{Jointly learning to recommend and advertise}. In
  \bibinfo{booktitle}{\emph{Proceedings of the 26th ACM SIGKDD International
  Conference on Knowledge Discovery \& Data Mining}}.
  \bibinfo{pages}{3319--3327}.
\newblock


\bibitem[\protect\citeauthoryear{Zhu, Jin, Tan, Pan, Zeng, Li, and Gai}{Zhu
  et~al\mbox{.}}{2017}]%
        {zhu2017optimized}
\bibfield{author}{\bibinfo{person}{Han Zhu}, \bibinfo{person}{Junqi Jin},
  \bibinfo{person}{Chang Tan}, \bibinfo{person}{Fei Pan},
  \bibinfo{person}{Yifan Zeng}, \bibinfo{person}{Han Li}, {and}
  \bibinfo{person}{Kun Gai}.} \bibinfo{year}{2017}\natexlab{}.
\newblock \showarticletitle{Optimized cost per click in taobao display
  advertising}. In \bibinfo{booktitle}{\emph{Proceedings of the 23rd ACM SIGKDD
  International Conference on Knowledge Discovery and Data Mining}}.
  \bibinfo{pages}{2191--2200}.
\newblock


\end{thebibliography}
	
\end{document}